\documentclass[10pt,twocolumn,letterpaper]{article}

\usepackage[pagenumbers]{cvpr} 

\usepackage{bm}
\usepackage{amsmath}
\usepackage{bbm}
\usepackage{float}

\usepackage[dvipsnames]{xcolor}
\newcommand\lft{\mathopen{}\left}
\newcommand\rgt{\aftergroup\mathclose\aftergroup{\aftergroup}\right}

\DeclareMathOperator*{\argmin}{\arg\min}

\definecolor{cvprblue}{rgb}{0.21,0.49,0.74}
\usepackage[pagebackref,breaklinks,colorlinks,citecolor=cvprblue]{hyperref}
\usepackage[capitalize]{cleveref}
\usepackage{url}

\def\paperID{1639} 
\def\confName{CVPR} 
\def\confYear{2024}

\title{OHTA: One-shot Hand Avatar via Data-driven Implicit Priors}

\author {
    Xiaozheng Zheng  \footnotemark[1] \quad 
    Chao Wen         \footnotemark[1] \quad  
    Zhuo Su                           \quad 
    Zeran Xu                          \quad 
    Zhaohu Li                         \quad 
    Yang Zhao                         \quad 
    Zhou Xue         \footnotemark[2]\\
    PICO, ByteDance
}

\begin{document}
\twocolumn[{%
\renewcommand\twocolumn[1][]{#1}%
\maketitle

\begin{center}
    \vspace{-1.5em}
    \centering
    \captionsetup{type=figure}
    \includegraphics[width=0.98\textwidth ]{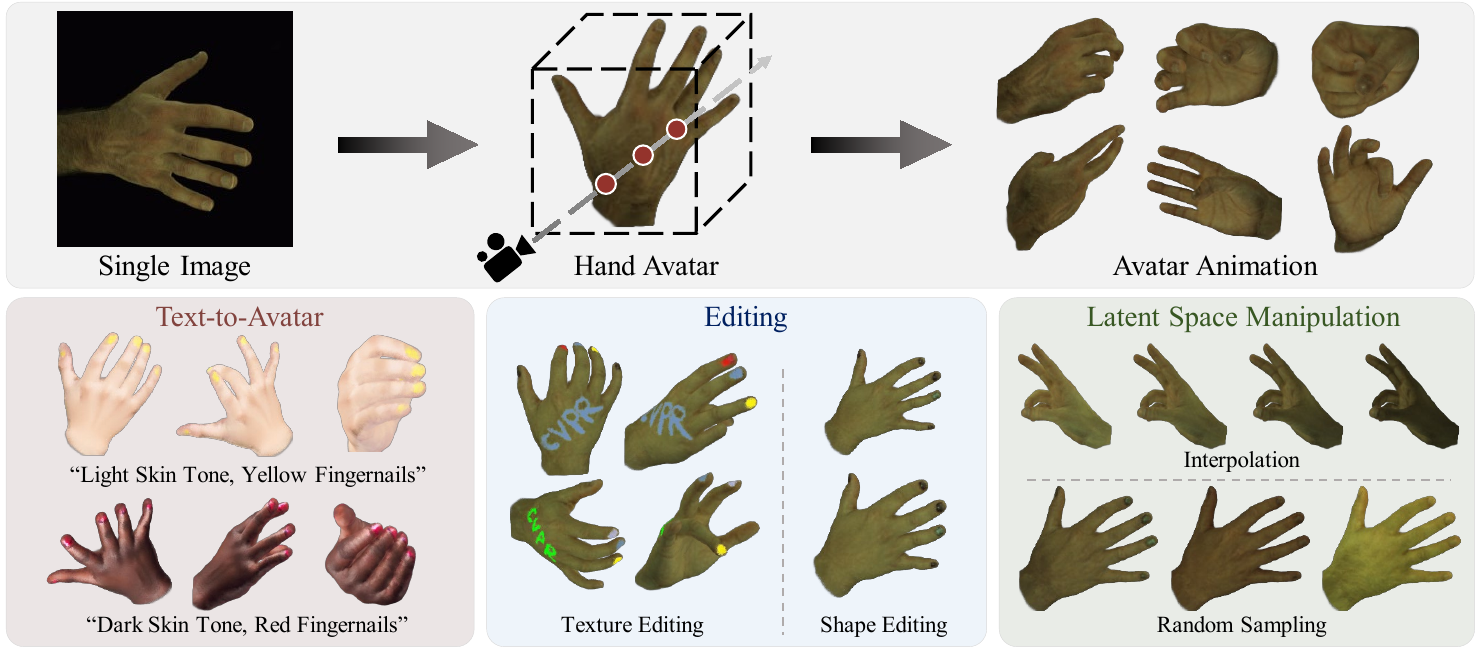}
        \captionof{figure}{We introduce a novel approach capable of creating implicit animatable hand avatars using just a single image. Our framework facilitates 1) text-to-avatar conversion, 2) hand texture and geometry editing, and 3) interpolation and sampling within the latent space.}
    \label{fig:teaser}
\end{center}%
}]
\maketitle

{
  \renewcommand{\thefootnote}%
    {\fnsymbol{footnote}}
  \footnotetext{Project page: \url{https://zxz267.github.io/OHTA}} 
}

{
  \renewcommand{\thefootnote}%
    {\fnsymbol{footnote}}
  \footnotetext{$^{*}$ Equal contribution, $^{\dagger}$ Corresponding author} 
}

\begin{abstract}
In this paper, we delve into the creation of one-shot hand avatars, attaining high-fidelity and drivable hand representations swiftly from a single image.
With the burgeoning domains of the digital human, the need for quick and personalized hand avatar creation has become increasingly critical. 
Existing techniques typically require extensive input data and may prove cumbersome or even impractical in certain scenarios.
To enhance accessibility, we present a novel method OHTA (\textbf{O}ne-shot \textbf{H}and ava\textbf{TA}r) that enables the creation of detailed hand avatars from merely one image.
OHTA tackles the inherent difficulties of this data-limited problem by learning and utilizing data-driven hand priors.
Specifically, we design a hand prior model initially employed for 1) learning various hand priors with available data and subsequently for 2) the inversion and fitting of the target identity with prior knowledge.
OHTA demonstrates the capability to create high-fidelity hand avatars with consistent animatable quality, solely relying on a single image. Furthermore, we illustrate the versatility of OHTA through diverse applications, encompassing text-to-avatar conversion, hand editing, and identity latent space manipulation.
\end{abstract}

\vspace{-1em}
\section{Introduction}
\label{sec:intro}

The progress in digital humans is reshaping our everyday lives by blending the physical and digital realms. 
In this emerging reality, human hands have a central role in creating a close and interactive experience, acting as the primary way people connect with the digital world. 
Consequently, it is vital to convert the hands of users into digital forms, allowing for the creation of personalized, controllable, and highly realistic representations in the virtual environment.

Conventional approaches of hand appearance modeling have typically relied on texture maps and colored meshes~\cite{qian2020html,potamias2023handy,gao2022cyclehand,chen2021model,li2022nimble,gao2022dart}.
However, the construction of personalized hand meshes and texture maps often demands the use of costly scanning data and artistic expertise. 
Recent efforts focus on creating hand avatars using data-driven approaches \cite{corona2022lisa,chen2023hand,karunratanakul2023harp,mundra2023livehand,guo2023handnerf,qu2023novel}. 
Many of these studies have concentrated on creating neural implicit hand avatars, as this form has demonstrated the ability to effectively model high-fidelity avatars across various human body parts~\cite{corona2022lisa,jiang2022selfrecon,weng2022humannerf,chen2023hand,mundra2023livehand,buhler2023preface}.
Despite the notable accomplishments in recent implicit hand representation, a noteworthy limitation persists, which entails the necessity of employing extensive sequential or multi-view images to procure a high-fidelity animatable hand avatar. 
This requirement poses a considerable challenge for general users, as the acquisition of such dense input data is often labor-intensive and, in certain cases, may not be practical. 
To broaden the scope of personalized hand avatar creation to more diverse situations, we aim to propose a novel approach capable of one-shot reconstruction of high-fidelity hand avatars.
This endeavor presents a substantial challenge, given that it places personalized hand avatar creation at a crossroads, necessitating a data-driven method despite the scarcity of available data.

In response to this challenge, we introduce OHTA (\textbf{O}ne-shot \textbf{H}and ava\textbf{TA}r) which captures data-driven priors to facilitate one-shot high-fidelity hand avatars creation. 
Our motivation stems from the fact that leveraging priors can complement unobserved information due to the inherent similarities shared by diverse human hands.  
To this end, we decouple our framework into two stages: 1) hand prior knowledge learning, and 2) one-shot reconstruction with the aid of prior knowledge.
In the first stage, we employ the Hand Prior Network (HPNet) to capture prior knowledge from the accessible data with multiple identities.
In the second stage, we utilize HPNet to inverse a similar hand and subsequently optimize it to the target identity with prior knowledge for regularization, based on the given image.

HPNet plays a pivotal role since its capacity for both 1) \emph{prior transferring} and 2) \emph{prior learning} significantly impacts the ultimate one-shot performance.
Therefore, designing such a network is non-trivial and presents a dual challenge: the architecture must be well-suited for two purposes simultaneously.
To address the challenges, we adopt corresponding designs in HPNet.
For \emph{prior transferring}, we design HPNet in a mesh-guided manner since the aid of the mesh enables the learned prior knowledge to transfer to novel identities.
Both geometry and texture priors learning of HPNet are guided by the mesh information, achieving robustness towards novel poses and identities.
For \emph{prior learning}, we equip HPNet with 1) the multi-resolution field based on the mesh scaffold and 2) identity-specific albedo and identity-shared shadow disentanglement.
The multi-resolution field facilitates learning detailed texture prior, and the disentanglement models transferable self-occlusion effect, both of which contribute to high-fidelity one-shot hand avatars.
With all these designs, OHTA is capable of addressing many downstream tasks (as shown in \cref{fig:teaser}), including 1) hand avatar creation using text prompts, 2) editing of hand geometry and appearance, and 3) appearance interpolation and sampling in identity latent space.

In summary, our contributions can be listed as follows:
\begin{itemize}
    \item We introduce the first framework for one-shot implicit hand avatar creation. 
    These avatars exhibit high-fidelity and consistent animatable quality.
    \item We present the Hand Prior Network exploiting transferable geometry, albedo, and shadow priors, contributing to the high fidelity of one-shot hand avatars.
    \item We substantiate the efficacy and robustness of our framework through a comprehensive series of experiments and showcase its utility in diverse applications, including text-to-avatar, hand editing, and latent space manipulation.
\end{itemize}

\section{Related Work}
\label{sec:related}

\noindent
\textbf{Animatable Hand Avatar.}
Research in creating personalized, animatable hand avatars has advanced through two main approaches. 
Some works utilize explicit geometry and texture.
HTML \cite{qian2020html} is prominent for its texture inference on the MANO model \cite{romero2017embodied}.
NIMBLE \cite{li2022nimble} extends the hand skeleton model \cite{li21piano}, capable of inferring surface with texture maps.
Handy \cite{potamias2023handy} improves realism using a GAN-based texture model \cite{karras2021alias}, but these methods often suffer from poor adaptability due to limited training data.
HARP \cite{karunratanakul2023harp} addresses this by adjusting geometry and texture for individual identities.
Another line of research explores neural implicit representations for more realistic texture modeling.
LISA \cite{corona2022lisa} learns an implicit color and surface representation. 
HandAvatar \cite{chen2023hand} employs disentangled implicit representations for hand geometry, albedo, and illumination.
HandNeRF \cite{guo2023handnerf} and LiveHand \cite{mundra2023livehand} extend NeRF \cite{mildenhall2021nerf} techniques with deformation fields for hands, even achieving real-time performance. 
HO-NeRF \cite{qu2023novel} further explores hand-object reconstruction using implicit avatars. 
Notwithstanding their accomplishments, it is worth noting that these models are unable to create high-fidelity hand avatars from a single image, as is achievable with our proposed approach.

\setlength{\belowcaptionskip}{-5pt}
\begin{figure*}
    \centering
    \includegraphics[width=0.96\textwidth]{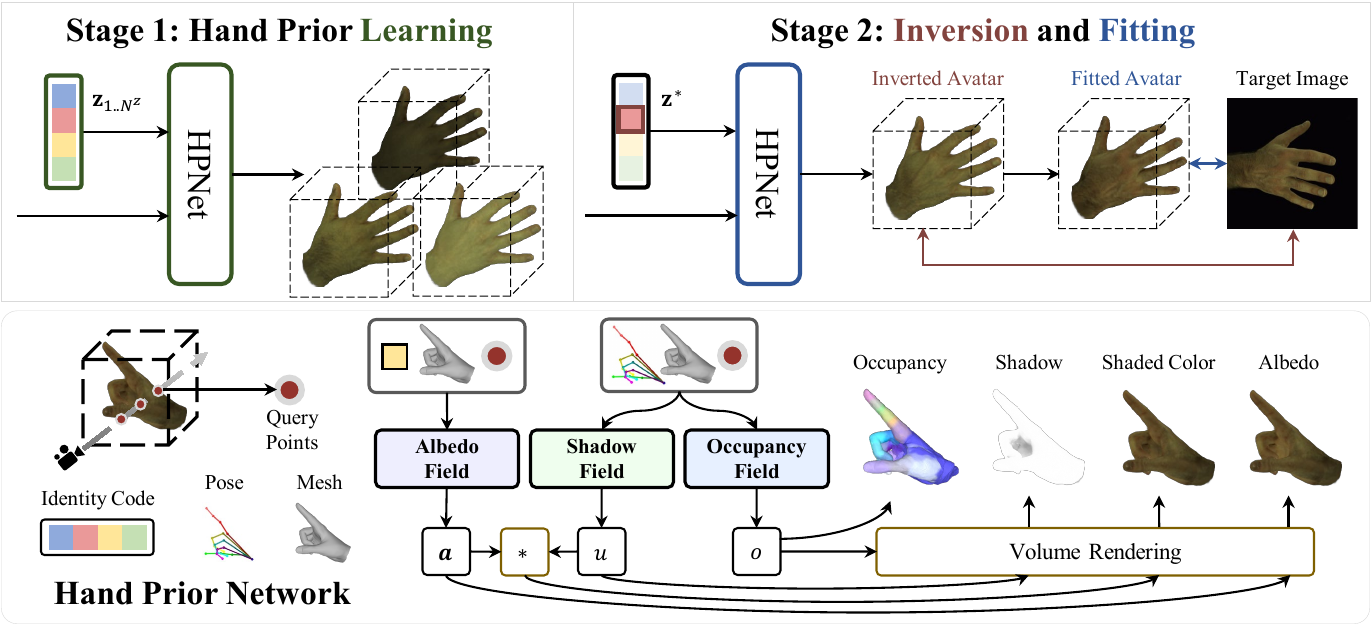}
    \caption{
    The two-stage framework of OHTA (above) and the Hand Prior Network (below). 
    For stage-1, OHTA optimizes \textcolor{OliveGreen}{identity code} and \textcolor{OliveGreen}{HPNet} to capture various hand priors.
    For stage-2, OHTA first optimizes \textcolor{Mahogany}{identity code} for texture inversion, then optimizes \textcolor{Blue}{HPNet} for texture fitting to capture the details.
    HPNet consists of three fields (\ie albedo, shadow, and occupancy) for capturing transferable hand prior knowledge.
    Combined with the albedo, shadow, and occupancy values, we use volume rendering to obtain final shaded color images.
    }
    \label{fig:whole-pipeline}
    \vspace{-1em}
\end{figure*}

\noindent
\textbf{One-shot Human Avatar.}
A similar task to one-shot hand avatar modeling is reconstructing a human body avatar from a single image, with methods divided into ``explicit'' and ``implicit'' categories. 
Explicit methods \cite{keepitSMPL, HMR18, joo2021exemplar, kolotouros2019learning, pavlakos2019expressive, Kocabas_2021_ICCV, alldieck2019learning, alldieck2019tex2shape, lazova2019360, ma2020learning, zhu2019detailed} rely on parametric models \cite{loper2023smpl, pavlakos2019expressive, STAR_ECCV2020} to estimate a minimally-clothed body or to regress body shape offsets, often struggling with loose clothing representation. 
DINAR \cite{svitov2023dinar} integrates neural textures with the explicit model for enhanced photo-realism.
Implicit methods, using topology-agnostic fields, aim for higher realism. 
Many works \cite{saito2019pifu,saito2020pifuhd, he2020geo, zheng2021pamir} utilize large datasets to estimate body avatars with pixel-aligned features, while others~\cite{huang2020arch,he2021arch++} further propose avatar animation. 
However, these approaches are often constrained by training data, limiting their generalizability.
NeRF-based methods \cite{weng2022humannerf, xu2021h, su2021nerf, peng2021animatable, zhao2022humannerf, jiang2023instant} typically require dense inputs but have evolved to accommodate few-shot or one-shot scenarios by incorporating various priors. 
ActorsNeRF \cite{mu2023actorsnerf} adapts to new subjects by pre-training category-level human NeRFs, while ELICIT \cite{huang2022elicit} and SHERF \cite{hu2023sherf} employ semantic priors or 3D-aware features for one-shot avatar creation.
For faces, 
CG-NeRF \cite{jo2023cg} supports one-shot 3D-aware face synthesis with a conditional NeRF.
PhoneScan \cite{cao2022authentic} learns identity and expression prior with CNNs from large-scale data, and finetunes the prior model for few-show head avatar creation.
Preface \cite{buhler2023preface} also trains an identity-conditioned prior model and achieves few-shot high-fidelity 3D head creations with its inversion and fitting.
Compared with PhoneScan and Preface, our method further exploits geometry, albedo, and shadow priors to create one-shot animatable hand avatars with consistent driving performance.

\noindent
\textbf{Mesh-guided Representation.}
Parametric models generate meshes from pose and shape parameters \cite{li2017learning,pavlakos2019expressive,romero2017embodied,loper2023smpl}, but these meshes have unalterable topology structure. 
Implicit representations are increasingly researched to address these limitations, with studies focusing on human \cite{deng2020nasa,mihajlovic2021leap,tiwari2021neural,saito2021scanimate,wang2021metaavatar,chen2021snarf,mihajlovic2022coap,alldieck2021imghum} and hand geometry \cite{karunratanakul2021skeleton,chen2023hand,huang2023phrit}, often using rigid transformations or skeletons for prediction. 
COAP \cite{mihajlovic2022coap} and PairOF \cite{chen2023hand} enhance robustness by incorporating mesh information and localized encoders or part-pair-wise decoders, respectively.
Mesh information also aids implicit texture modeling \cite{peng2021neural,lombardi2021mixture,zheng2022structured,yang2022neumesh,wu2023nerf,chen2023hand,bai2023learning}. 
NeuMesh \cite{yang2022neumesh} and DE-NeRF \cite{wu2023nerf} place features on vertices for editable radiance fields.
Recent implicit human body \cite{peng2021neural,zheng2022structured} and face \cite{grassal2022neural, bai2023learning} modeling also rely on mesh information for guidance.
HandAvatar \cite{chen2023hand} uses barycentric sampled anchor upon meshes for hand modeling. 
Our method differs by employing multi-resolution fields on the mesh scaffold for enhanced robustness and fidelity.

\noindent
\textbf{Single Image to 3D.}
Recently, there has been a surge in research focused on learning how to create 3D models from a single image
~\cite{mescheder2019occupancy, yu2021pixelnerf}.
Zero-1-to-3~\cite{liu2023zero} pioneers the zero-shot single-image to 3D conversion. 
Recent works like One-2-3-45 \cite{liu2023one}, Magic123 \cite{qian2023magic123}, and Consistent123 \cite{lin2023consistent123} build upon Zero-1-to-3 to obtain more 3D-consistent results by introducing more priors.
Although such methods have pre-trained networks on large datasets, there is an underutilization of specific domain knowledge.
In contrast to these methods, we are dedicated to considering the distinctive properties of the hand to create animatable avatars.

\setlength{\abovedisplayskip}{3pt} 
\setlength{\belowdisplayskip}{3pt}
\setlength{\abovedisplayshortskip}{3pt}
\setlength{\belowdisplayshortskip}{3pt}

\section{Method}
\label{sec:method}
In this section, we provide a detailed description of the proposed hand representation and complete avatar reconstruction pipeline (\cref{fig:whole-pipeline}).
We first introduce a novel hand representation that encodes multi-resolution neural implicit fields at a mesh scaffold~(\cref{sec:hand model}).
As for the one-shot hand avatar creation pipeline, we decouple it into two stages.
In the first stage~(\cref{sec:HPNet}), we devise an effective training scheme that captures hand prior knowledge through the HPNet.
In the second stage~(\cref{sec:one-shot fitting}), we invert and optimize the HPNet to create high-quality hand avatars according to a single input image.

\subsection{OHTA Hand Representation}
\label{sec:hand model}
Our representation of the hand comprises implicit geometry and neural texture fields based on a mesh scaffold.

\noindent
\textbf{Mesh Scaffold.}
We adopt MANO-HD \cite{chen2023hand}, a super-resolution variant of MANO \cite{romero2017embodied}, as the mesh scaffold.
(1)~For training HPNet, we use personalized MANO-HD by leveraging a multi-layer perceptron (MLP) $\mathcal{M}_{shape}$ to refine the template mesh. 
The refined template mesh is defined as $\mathbf{M}^{+} =\bar{\mathbf M}+\mathcal M_{shape}([\mathcal{P}(\bar{\mathbf M}), \bm{\theta}, \mathbf{z}])$, where $\bar{\mathbf M}$, $\mathcal{P}$, $\bm{\theta}$, $\mathbf{z}$ and $[\cdot]$ denote the template mesh, positional encoding, pose parameters, identity code and concatenation respectively.
The MLP can be optimized via IoU loss $\mathcal{L}_{shape}$, detailed in \cite{liu2019soft}.
The template mesh can be deformed to posed mesh with a Linear Blend Skinning transformation according to pose parameters.
(2)~For one-shot hand avatar reconstruction, where acquiring a precisely refined mesh from a single view is unfeasible, we use a mesh $\mathbf{M}$ represented by shape $\bm{\beta}$ and pose $\bm{\theta}$ parameters as the input.
For simplicity, the input mesh is denoted as $\mathbf{M}$ in the following text.

\noindent
\textbf{Geometry Network.}
To achieve robust geometry modeling towards novel identities and poses, we utilize PairOF \cite{chen2023hand} to predict occupancy values.
Given $N^{q}$ query points $ \mathbf{q} \in \mathbb{R}^{N^{q}\times3}$, hand mesh and hand pose parameters, PairOF $f^{\mathcal{O}}$ predicts the occupancy value $o_{\mathbf{q}} = f^{\mathcal{O}}(\mathbf{q}, \mathbf{M}, \bm{\theta})$ to indicate whether the point locates inside or out of the surface. Hence, the hand surface can be formulated as $\{ \mathbf{q} | o_{\mathbf{q}} = 0.5 \}$.
For more details, please refer to \cite{chen2023hand,mihajlovic2022coap}.

\begin{figure}[t]
    \centering
    \includegraphics[width=0.47\textwidth ]{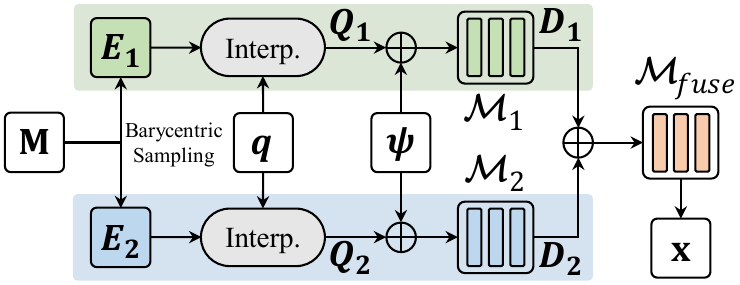}
    \caption{Multi-resolution Field. For simplicity, we take two resolutions for example. $\oplus$ denotes feature concatenation.}
    \label{fig:multi-resolution field}
\end{figure}

\noindent
\textbf{Multi-resolution Field.}
To achieve high-fidelity texture modeling, we designed a Multi-resolution Field~(\cref{fig:multi-resolution field}) with the mesh guidance.
More specifically, we uniformly sample $N^{p}_{k}$ points $\mathbf{P}_{k}$ on the input mesh and represent them with barycentric coordinates for each resolution, where $k \in \{1, ..., K\}$ denotes the resolution level. 
For each point in $\mathbf{P}_{k}$, we attach them with a learnable point encoding of $N^c$ dimension that is randomly initialized.
In this way, we can obtain point encodings $\mathbf{E}_{k}$ of a resolution.
For query points $\mathbf{q}$, we conduct spatial interpolation to acquire the queried encoding of this resolution $\mathbf{Q}_{k}=\text{interp}(\mathbf{q}, \mathbf{E}_{k})$.
Specifically, we first extract the encodings of $N^{n}$ neighbor points $\mathcal{K}(\mathbf{E}_{k}) \in \mathbb{R}^{N^{n} \times N^{q} \times N^c}$ of $\mathbf{P}_{k}$, where $\mathcal{K}$ denotes \textit{k}-nearest neighbors. 
Then, we perform a weighted average with the inverse Euclidean distances of those points to the query point as the weights to acquire the queried encoding of this resolution $\mathbf{Q}_{k} \in \mathbb{R}^{N^{q} \times N^{c}}$.
After that, we feed queried encodings with the field-specific information $\mathbf{\bm{\psi}}$ to resolution-specific MLP $\mathcal{M}_{k}$ to obtain resolution-specific features:
$\mathbf{D}_{\mathbf k} = \mathcal{M}_{k}(\mathbf{Q}_{k}, \mathbf{\bm{\psi}})$, where $\mathbf{D}_{\mathbf k} \in \mathbb{R}^{N^{q} \times N^{d}}$,
and $\mathbf{\bm{\psi}}$ is designed to distinguish the albedo-specific field $f^{\mathcal{A}}$ and shadow-specific field $f^{\mathcal{U}}$, which are combined with the multi-resolution mechanism to better represent the texture, as described later.
Finally, we make use of another MLP $\mathcal{M}_{fuse}$ to fuse resolution-specific features of multi-resolutions to predict the target field value $\mathbf{x}_{\mathbf q} = \mathcal{M}_{fuse}([\mathbf{D}_{1}\, ..., \mathbf{D}_{K}])$.
\noindent
The overall hand representation comprises the above components, as defined formally:
\begin{equation}\label{eq:mapping}
    \mathcal{H}: (\mathbf{q}, \mathbf{M}, \bm{\theta}, \textbf{z}; f^{\mathcal{A}}, f^{\mathcal{U}}, f^{\mathcal{O}}) \mapsto (o, \bm{c})\;,
\end{equation}
where $o$ and $\bm{c}$ denote the occupancy and shaded color.

\setlength{\belowcaptionskip}{-5pt}

\subsection{Hand Prior Learning}
\label{sec:HPNet}
To exploit the potential of OHTA representation for one-shot reconstruction, learning transferrable and abundant hand prior knowledge is of priority.
Inspired by~\cite{cao2022authentic}, we propose a novel paradigm for encoding those prior to the HPNet through a set of accessible data from multiple identities.
On the dataset including $N^{z}$ identities, our goal is to learn multiple hand avatars within a network, with the respective identity codes $\mathbf{z}_{1\dots N^{z}}$ and network parameters optimized.

\noindent
\textbf{Geometry Prior.}
For the hand geometry prior learning, we first pre-train the PairOF module (\ie occupancy field) $f^{\mathcal{O}}$. 
We utilize data from different hand shapes to make HPNet learn the geometric priors of the hand which can transfer to new identities. 
With hand meshes derived from input $\mathbf{M}$, we sample point clouds with ${N^o}$ points as training data~\cite{mihajlovic2022coap}.

\noindent
\textbf{Texture prior.}
Subsequently, we focus on learning the texture prior knowledge of the hand.
We observe that the albedo features are specific to each identity, while the shadows, which arise due to the hand's self-occlusion, are common across different identities.
Hence, we disentangle the texture prior learning into albedo and shadow prior learning.
Our methodology leverages a multi-resolution field representation for simultaneous learning of both albedo and shadow fields by incorporating field-specific information $\mathbf{\bm{\psi}}$. 

\textit{Identity-specific Albedo Field.}
Since the albedo of the hands is dependent on the identity other than pose etc., we set $\mathbf{\bm{\psi}} = \mathbf{z}$ for identity-specific albedo field modeling.
For query points $\mathbf{q}$, we obtain the albedo $ \mathbf{a}_{\mathbf q}=f^{\mathcal{A}}(\mathbf{q}, \mathbf{M}, \bm{z})$.

\textit{Identity-shared Shadow Field.}
Modeling identity-specific environmental lighting conditions of hands like HandAvatar~\cite{chen2023hand} cannot capture shadow prior knowledge that can generalize to one-shot reconstruction.
To overcome this problem, our approach aims to model a more universal shadow prior, grounded in the observation that shadows from self-occlusion are largely consistent across different hands when performing identical poses. 
Hence, we build the shadow field with $\mathbf{\bm{\psi}} = \bm{\theta}$, where $\bm{\theta}$ is pose parameters without global rotations.
For query points $\mathbf{q}$, we obtain the shadow $ u_{\mathbf q}=f^{\mathcal{U}}(\mathbf{q}, \mathbf{M}, \bm{\theta})$.

Combining the albedo field and shadow field, we obtain a complete Multi-resolution Texture Field (MTField).
For query points $\mathbf{q}$, we derive the shaded color value $ \mathbf{c}_{\mathbf q} = u_{\mathbf q}\mathbf{a}_{\mathbf q} \in \mathbb{R}^{N^{q}\times 3}$.
For a casted camera ray $\mathbf{r}$, we uniformly sample $N^{q}$ queries $\{\mathbf q_i\}_{i=1}^{N^q}$, each of which has occupancy $o$ and texture $\mathbf{c}$. We render each pixel $\mathbf{I_{\mathbf{r}}}$ of image via volume rendering~\cite{mildenhall2021nerf}:
\begin{equation}
\label{eq:render}
\begin{array}{l}
\mathbf{I_{\mathbf{r}}} =\sum\limits_{i=1}^{N^q}\lft(\prod\limits_{j=1}^{i-1}(1-o_{\mathbf q_j})\rgt) o_{\mathbf q_i} \mathbf{c}_{\mathbf{q}_{i}}.
\end{array}
\end{equation}

\noindent
\textbf{Optimization.}
The training of HPNet consists of two steps.
In the first step, we pre-train $f^{\mathcal{O}}$ with the MANO pose parameters annotations and shape parameters randomly sampled from a given range. 
The objective for optimizing PairOF is to minimize the error between ground truth $o^*$ and predicted occupancy values, \ie, $\mathcal L_{occ}=\frac{1}{N^o}\sum_{\mathbf q} (o_{\mathbf q} - o_{\mathbf q}^*)^2$. 
In the second step, we conduct end-to-end training for the mesh scaffold, occupancy field, and MTField.
Specifically, MTField is optimized with shadow regularization $\mathcal{L}_{shadow} =  \| \mathbf{U} - \mathbbm{1} \|_{1}$ and the reconstruction loss function (consisting of LPIPS loss \cite{zhang2018unreasonable} and $l_{1}$ loss):
\begin{equation}
\begin{array}{l}\label{eq:reconstruction}
\mathcal{L}_{rec} = \mathcal{L}_{\text{LPIPS}}(\mathbf{I}, \mathbf{I}^{*}) + \| \mathbf{I} - \mathbf{I}^{*} \|_{1},
\end{array}
\end{equation}
where $\mathbf{U}$, $\mathbf{I}$ and $\mathbf{I^{*}}$  denote the shadow values, the rendered image, and the ground truth image.
The shadow field is regularized to be close to 1 to ensure that it focuses more on the shadow caused by self-occlusion.
The full loss is a combination of the previous terms: $ \mathcal{L} = \mathcal{L}_{shape} + \mathcal{L}_{occ} + \mathcal{L}_{rec} + \lambda_{shadow} \mathcal{L}_{shadow}$, where $\lambda_{shadow}$ is used to balance different loss terms.

\subsection{One-shot Hand Avatar Reconstruction Pipeline}
\label{sec:one-shot fitting}
After hand prior learning, HPNet encodes various prior knowledge useful for one-shot reconstruction.
For each input image, we only optimize part of HPNet to reconstruct the target hand avatar.
The complete pipeline is as follows.

\noindent
\textbf{Preprocessing.}
We utilize the off-the-shelf hand pose estimator~\cite{ren2023decoupled} to predict the shape parameter $\bm{\beta}$, pose parameter $\bm{\theta}$, and camera pose of the input images.

\noindent
\textbf{Geometry.}
Since it is impractical to learn a refined mesh based on an image, we adopt meshes derived from hand shape parameters $\bm{\beta}$ as the input for occupancy prediction.

\begin{figure*}
    \centering
    \includegraphics[width=1.00\textwidth]{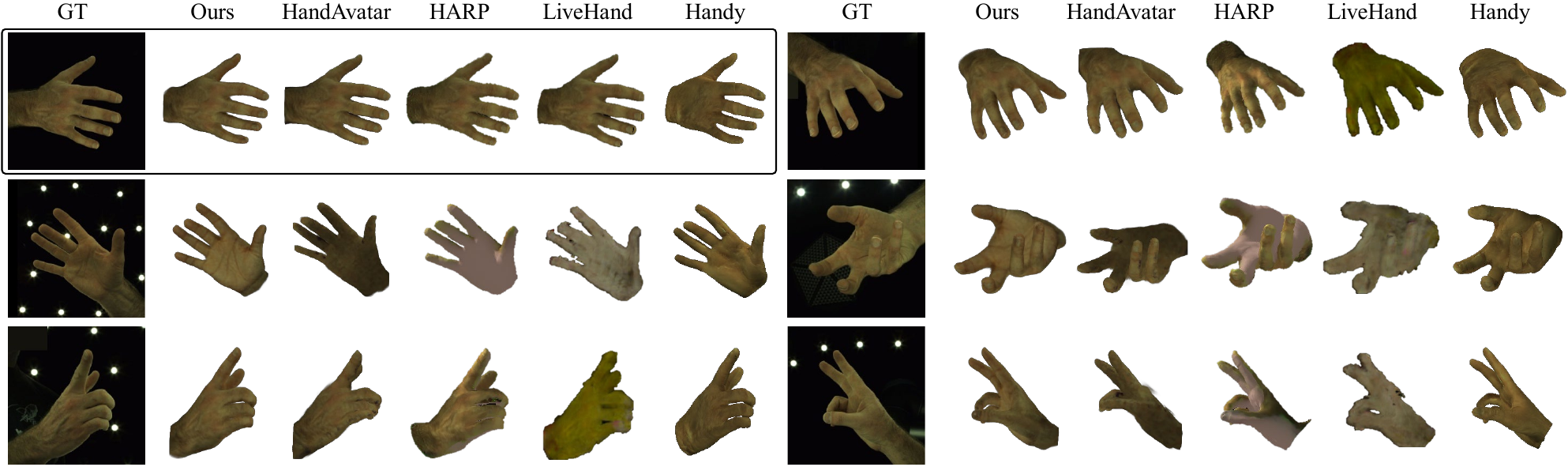}
    \caption{Qualitative comparison with state-of-the-art methods on InterHand2.6M~\cite{moon2020interhand2}. The black box indicates the input image.}
    \label{fig:interhand2.6m}
\end{figure*}

\noindent
\textbf{Texture Inversion.}
To provide good prior knowledge for the one-shot reconstruction, we first inverse a similar appearance from the MTField. 
More specifically, we solve an optimization problem to find identity code $\mathbf{z}^{*}$ and per-channel color calibration coefficients $\{ \mathbf{w}, \mathbf{b} \}$ that can produce a similar appearance to the target identity of the input image.
The color calibration alleviates the hand skin bias caused by the data used for prior learning, assisting better texture inversion for the target identity.
It works as follows: $\tilde{\mathbf{a}}_{\mathbf q} = \mathbf{w} * \mathbf{a}_{\mathbf q} + \mathbf{b}$.
During the inversion optimization, we keep the network weights frozen.
Given an input image $\mathbf{I}^{*}$ of the target identity, we optimize to render an image $\mathbf{I}$ that is similar to the target identity using our MTField with $\mathbf{z}^{*}$ and $\{ \mathbf{w}, \mathbf{b} \}$.
This procedure is optimized with the reconstruction loss function in \cref{eq:reconstruction}: $\argmin\limits_{\mathbf{z}^{*}, \mathbf{w}, \mathbf{b}} \mathcal{L}_{rec}$.

\noindent
\textbf{Texture Fitting.}
The goal of the texture fitting is to adapt the weights of the MTField to capture the details of the target identity from the input image.
We utilize prior knowledge in this procedure by two means. 
First, we only fine-tune the resolution-specific MLPs $\{\mathcal{M}_{k}\}_{k=1}^{K}$ and texture feature fusing MLP $\mathcal{M}_{fuse}$ of albedo field $f^{\mathcal{A}}$, while keeping other components (\eg, the point encodings and shadow field) frozen.
Hence, we can transfer the prior knowledge (\eg, shadow prior) well to the target avatar.
In this stage, all learnable parameters are denoted as $\xi$.
Second, we apply view regularization to avoid overfitting the target view. 
Specifically, we constraint the texture-fitting results of some reference views with different poses $\{\mathbf{R}_{i}\}_{i=1}^{N^{r}}$ to be close to the rendering results before texture-fitting $\{\tilde{\mathbf{R}}_{i}\}_{i=1}^{N^{r}}$, where $N^{r}$ is the number of the generated reference view.
With the prior knowledge, our reconstructed hand avatar can achieve stable animation while preserving the details of the target identity.
This fitting is conducted by minimizing a series of the reconstruction loss function in \cref{eq:reconstruction}:
\begin{equation}
\label{eq:inverse}
\argmin_{\xi} \mathcal{L}_{fit} = \mathcal{L}_{rec}(\mathbf{I}, \mathbf{I}^{*}) + \lambda_{ref} \sum_{i=1}^{N^{r}}\lft(\mathcal{L}_{rec}(\mathbf{R}_{i}, \tilde{\mathbf{R}}_{i})\rgt),
\end{equation}
where $\lambda_{ref}$ is used to balance different loss terms.

\section{Experiment}
\label{sec:experiment}
\subsection{Implementation Details and Metrics}
\label{sec:implementation}

\noindent
\textbf{Pre-training of PairOF.}
We adopt all right-hand annotations in InterHand2.6M \cite{moon2020interhand2} for pre-training like HandAvatar \cite{chen2023hand}.
To make the occupancy field more robust to new identities, we randomly sample the hand shape parameters $\bm{\beta}$ from a range $\pm3\sigma$ ensuring that the hand shapes are physically plausible.
We set $N^o=256$ for training with $\mathcal L_{occ}$.

\noindent
\textbf{End-to-end Prior Learning of HPNet.}
We utilized 21 subjects from the InterHand2.6M~\cite{moon2020interhand2} training set as the training data for HPNet. Specifically, each subject used one capture and left out 7 unseen poses for evaluation.
For the albedo field, we set $K=4$ and $N^{p}_{k}=\{512 \times 2^{k-1}\}_{k=1}^{4}$. 
For the shadow field, we set $K=1$ and $N^{p}_{1}=256$.
In addition, we set $N^n=4$, $N^q=64$, $N^c=128$, and $N^d=16$. 
The identity code $\mathbf{z}$ is learnable parameters initialized from a truncated normal distribution with standard deviation $\sigma=0.02$.
The rendering resolution is $256\times 256$.

\noindent
\textbf{One-shot Reconstruction.}
We use an off-the-shelf estimator~\cite{ren2023decoupled} to predict the MANO parameters of the in-the-wild images.
We set $N^r = 7$ and $\lambda_{ref}=0.2$. 
The reconstruction first takes $50$ steps for texture inversion and then $100$ steps for texture fitting (56 minutes with an A100 GPU).
For more details, please refer to our supplementary materials. 

\noindent
\textbf{Metrics.}
Consistent with previous works \cite{chen2023hand,corona2022lisa,mundra2023livehand}, we report LPIPS \cite{zhang2018unreasonable}, PSNR, and SSIM \cite{wang2004image} to reflect image similarity as the metrics of rendering quality.

\subsection{Evaluation of One-shot Avatar Reconstruction}
\label{sec:comparisons}
We comprehensively compare OHTA with previous methods~\cite{chen2023hand,karunratanakul2023harp,potamias2023handy,mundra2023livehand} under the one-shot reconstruction scenario with different data sources. 
Additionally, we benchmark against methods~\cite{weng2022humannerf,jiang2022selfrecon,chen2023hand} that employ a large number of images, aiming to gauge the potential performance ceiling in one-shot scenarios. 
Moreover, we conduct novel-view comparisons with LISA~\cite{corona2022lisa} and One-2-3-45~\cite{liu2023one}.

\begin{figure}
    \centering
    \includegraphics[width=0.48\textwidth]{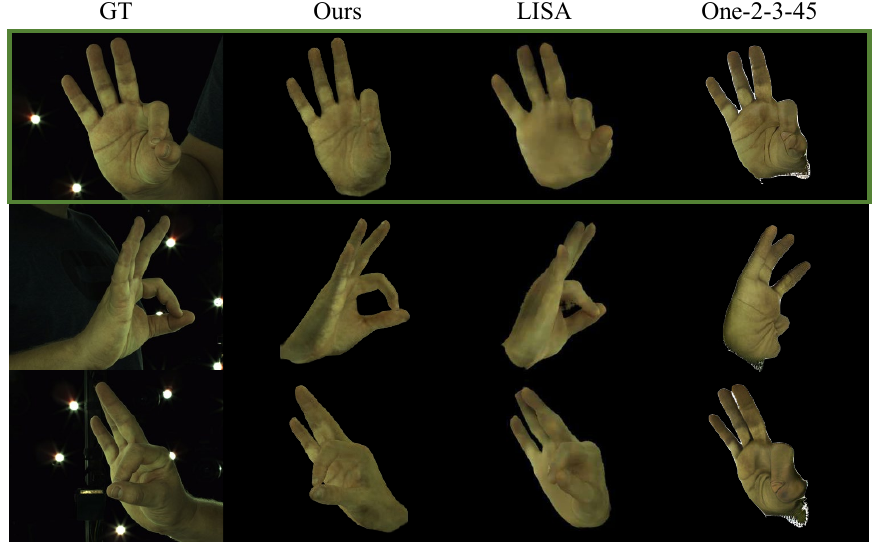}
    \caption{Comparison of novel view synthesis on InterHand2.6M~\cite{moon2020interhand2}. The green box indicates the input view.}
    \label{fig:novel-view}
\end{figure}

\begin{figure}
    \centering
    \includegraphics[width=0.48\textwidth]{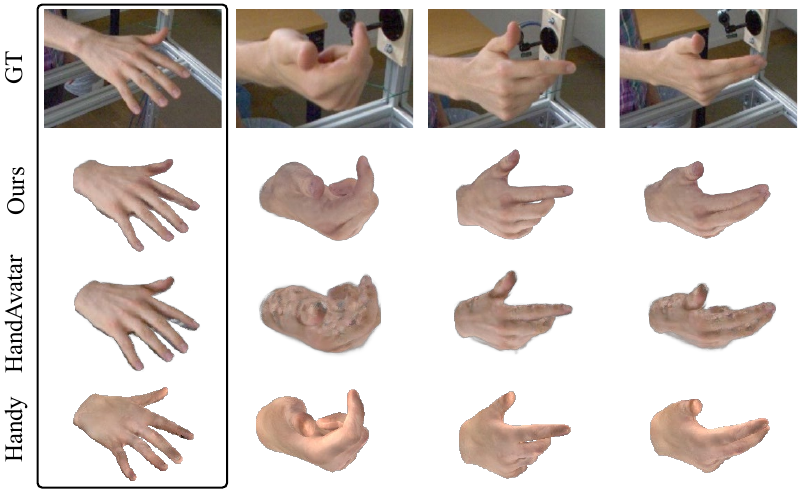}
    \caption{Qualitative comparison on HanCo~\cite{zimmermann2021contrastive}. }
    \label{fig:hanco}
\end{figure}

\noindent
\textbf{InterHand2.6M.}
We conduct evaluations on the testing set of InterHand2.6M \cite{moon2020interhand2}, following HandAvatar \cite{chen2023hand}.
The results of training on multi-view sequences are reported in \cite{chen2023hand}.
The results of training with a single image~(\ie one-shot) are based on their open-source codes \cite{chen2023hand,karunratanakul2023harp,potamias2023handy,mundra2023livehand}.
We do not update the hand pose parameters for HARP since the official implementation for a single image may collapse when updating them.
As for Handy \cite{potamias2023handy}, we begin by converting the MANO labels into Handy's format. We then optimize the latent vectors of the texture model in Handy and fit the single image using differentiable rendering.
As shown in \cref{tab:i26m-cap0}, our one-shot performance is significantly better than the existing methods \cite{karunratanakul2023harp,mundra2023livehand,chen2023hand,potamias2023handy}.
Note that \# denotes the number of images.
The qualitative comparisons in \cref{fig:interhand2.6m} demonstrate our robust performance for novel views and poses.
All the results demonstrate that data-driven priors can benefit the one-shot performance remarkably.
Meanwhile, our performance is comparable to that of methods using all the images for training, which further validates the quality of the avatars.

\cref{fig:novel-view} presents the novel view synthesis comparisons with LISA \cite{corona2022lisa} and One-2-3-45 \cite{liu2023one}.
Due to the unavailability of models, we use results reported in the original LISA paper for comparison.
Our one-shot performance can capture more details than LISA, owing to our adequate prior knowledge from the prior model.
Due to the underutilization of specific domain knowledge, the general 3D generation approach (One-2-3-45) cannot generate plausible novel views for animatable hand avatar creation.

\begin{table}
  \centering
  \resizebox{0.95\columnwidth}{!}{
  \begin{tabular}{cccccccc}
    \toprule 
     Method & \#Train & PSNR$\uparrow$ & LPIPS$\downarrow$ & SSIM$\uparrow$   \\
    \midrule
    SelfRecon   \cite{jiang2022selfrecon}    & 11,757       &  26.38  & 14.21 & 0.879   \\
    HumanNeRF   \cite{weng2022humannerf}     & 11,757       &  27.64  & 11.45 & 0.884   \\
    HandAvatar \cite{chen2023hand}           & 11,757       &  28.23  & 10.35 & 0.894   \\
    \midrule
    \midrule
    HandAvatar \cite{chen2023hand}           & 1  & 23.79  & 17.78  & 0.820    \\
    HARP     \cite{karunratanakul2023harp}   & 1  & 19.82  & 22.49  & 0.761    \\
    LiveHand \cite{mundra2023livehand}       & 1  & 23.01  & 18.64  & 0.763    \\
    Handy     \cite{potamias2023handy}       & 1  & 25.56  & 14.98  & 0.794    \\
    \midrule
     Ours          & 1      &  \textbf{26.11}  & \textbf{12.93} & \textbf{0.864}   \\
    \bottomrule
  \end{tabular}
  }
  \caption{Evaluation results on InterHand2.6M.}
  \label{tab:i26m-cap0}
\end{table}

\begin{table}
  \centering
  \resizebox{.76\columnwidth}{!}{
  \begin{tabular}{cccccccc}
    \toprule 
     Method & PSNR$\uparrow$ & LPIPS$\downarrow$ & SSIM$\uparrow$   \\
    \midrule
    HandAvatar \cite{chen2023hand}        & 19.76     & 14.41    & 0.846    \\
    Handy     \cite{potamias2023handy}    & 21.11     & 11.71    & 0.768    \\
    \midrule
     Ours                                 & \textbf{22.15}     & \textbf{11.55}    & \textbf{0.886}    \\
    \bottomrule
  \end{tabular}
  }
  \caption{Evaluation results on HanCo.}
  \label{tab:hanco}
\end{table}

\noindent
\textbf{HanCo.}
To show OHTA is robust for the appearance different from the training data of the HPNet, we also compare with existing methods on the HanCo~\cite{zimmermann2021contrastive} dataset.
We take HandAvatar \cite{chen2023hand} and Handy \cite{potamias2023handy} for comparisons since they show robust performance for one-shot reconstruction on InterHand2.6M (\cref{tab:i26m-cap0}).
We take 1 sequence of 4 cameras for the experiment, with 1 frame for training and others for testing.
For more details, please refer to our supplementary materials.
As shown in \cref{tab:hanco}, our method outperforms other methods consistently in all metrics.
We also show the qualitative comparisons in \cref{fig:hanco}, demonstrating our method can obtain consistent performance for free-pose animation.

\begin{figure}
    \centering
    \includegraphics[width=0.48\textwidth]{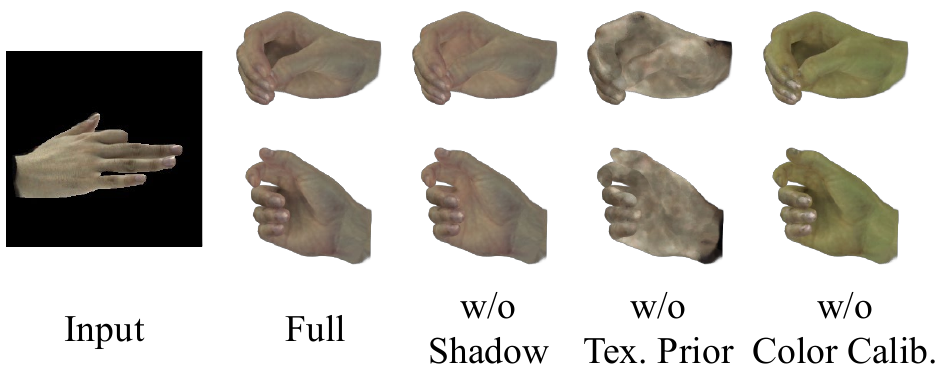}
    \caption{Visual results for ablation study with in-the-wild input.}
    \label{fig:ablation}
\end{figure}

\begin{figure}
    \centering
    \includegraphics[width=0.48\textwidth]{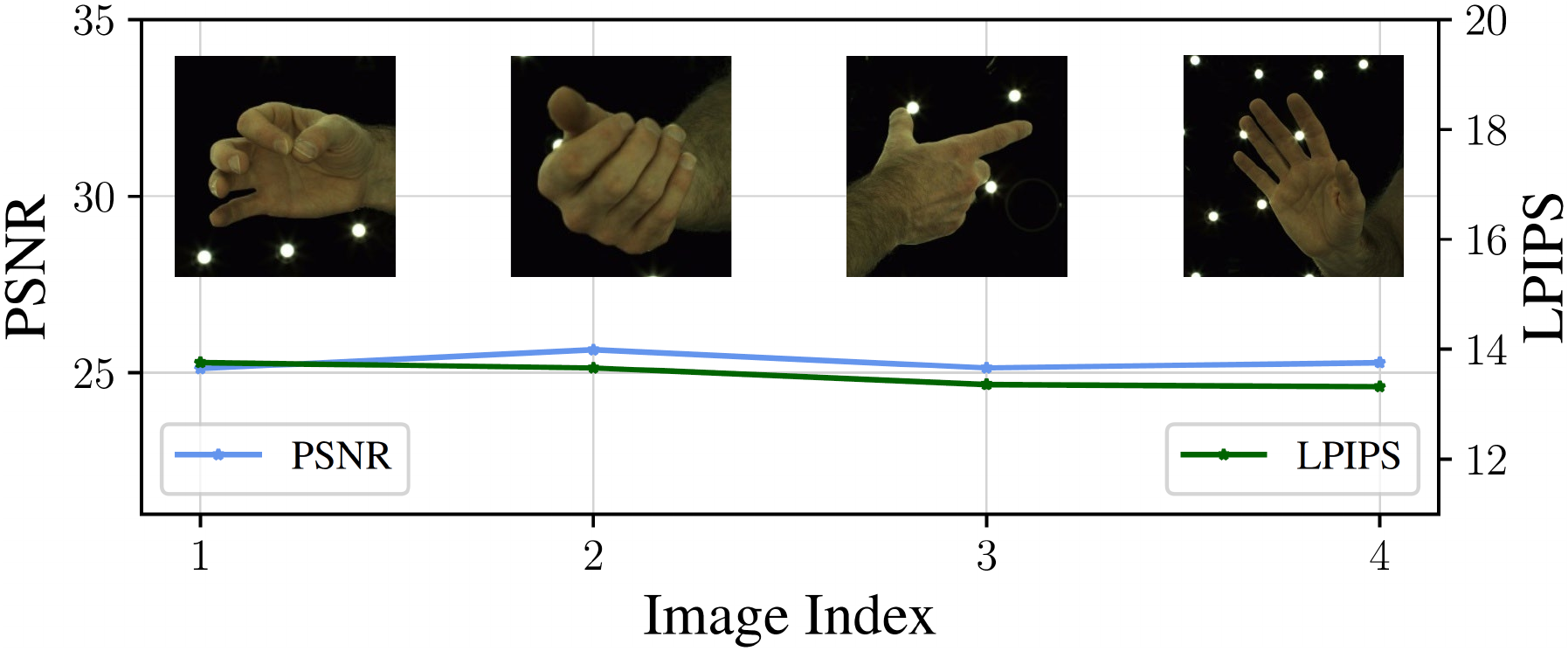}
    \caption{Robustness towards different input images. The images above the figure are the corresponding input images.}
    \label{fig:model-analysis-robustness-input}
    \vspace{-0.1cm}
\end{figure}

\begin{figure*}
    \centering
    \includegraphics[width=1.00\textwidth]{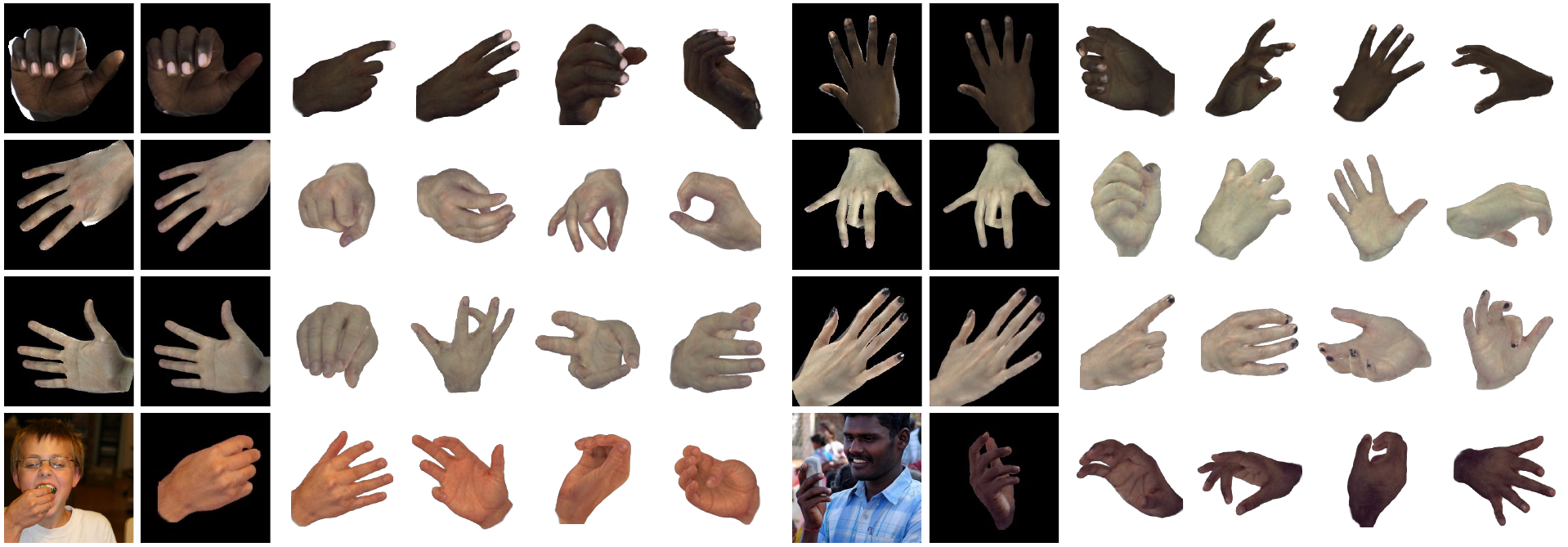}
    \caption{In-the-wild results from real-captured images and MSCOCO~\cite{lin2014microsoft} dataset. 
    The last line shows results on MSCOCO.}
    \label{fig:model-analysis-in-the-wild}
    \vspace{-0.2cm}
\end{figure*}

\vspace{-5px}
\begin{figure}
    \centering
    \includegraphics[width=0.45\textwidth]{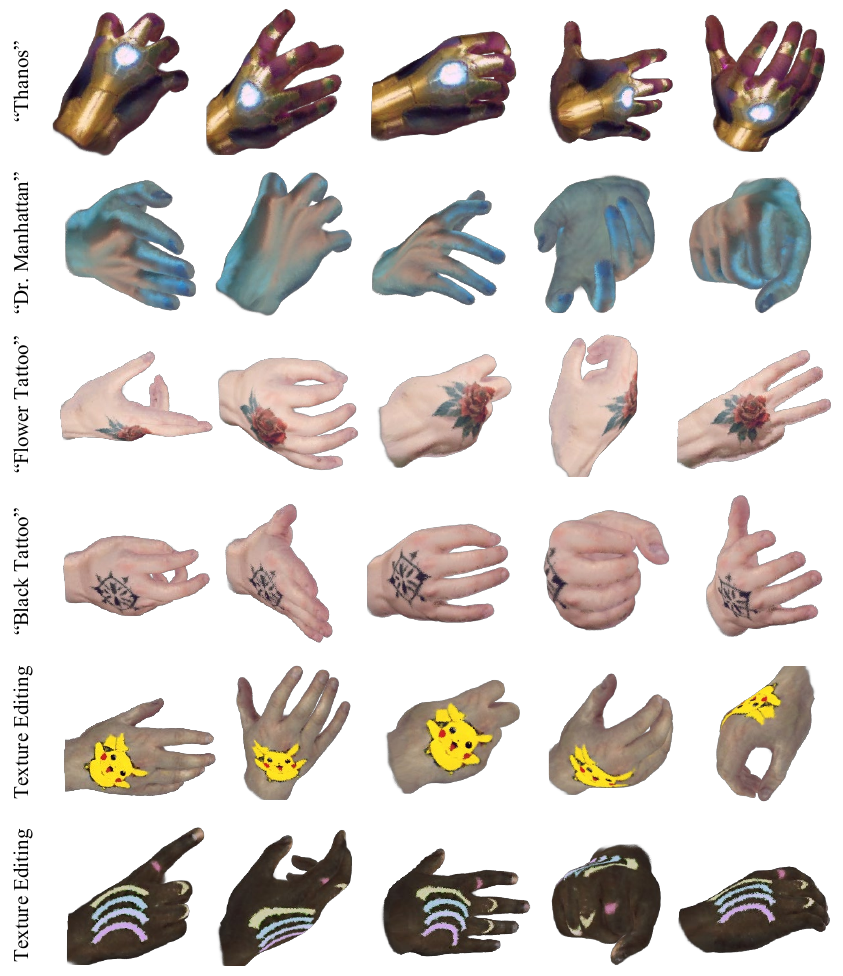}
    \caption{Text-to-avatar and texture editing. }
    \label{fig:application}
    \vspace{-0.4cm}
\end{figure}

\begin{table}[t]
  \centering    
  \resizebox{0.85\columnwidth}{!}{
  \begin{tabular}{lccccccc}
    \toprule 
     Method  & PSNR$\uparrow$ & LPIPS$\downarrow$ & SSIM$\uparrow$   \\
    \midrule
     Full Model                         & \textbf{27.64} & \textbf{12.23} & \textbf{0.896}     \\
     w/o Multi-resolution               & 27.11 & 12.96 & 0.890     \\
     w/o Shadow Field                   & 27.09 & 12.62 & 0.889     \\
     w/o Shape Fitting                  & 27.27 & 12.63 & 0.892     \\
    \bottomrule
  \end{tabular}
  }
  \caption{Ablation of HPNet for unseen poses.}
  \label{tab:ablation-HPNet}
\end{table}

\subsection{Ablation Study}
Our ablation studies contain two parts. 
One part is for justifying our components of HPNet.
The other part is to show our designs are effective for one-shot reconstruction.

\noindent
\textbf{Hand Prior Network.}
The evaluation is conducted in 7 unseen poses as described in \cref{sec:implementation}.
As shown in \cref{tab:ablation-HPNet}, employing multi-resolution fields contributes greatly to all the metrics.
Meanwhile, modeling pose-aware, identity-shared shadow is beneficial for overall performance, validating the existence of a shared self-occlusion effect for hands.
At last, we demonstrate learning identity-specific hand shape offset is significantly important when the data is abundant to remove the shape inaccuracy of MANO fitting.

\begin{table}[t!]
  \centering
  \resizebox{0.82\columnwidth}{!}{
  \begin{tabular}{lccccccc}
    \toprule 
     \# & Method  & PSNR$\uparrow$ & LPIPS$\downarrow$ & SSIM$\uparrow$   \\
    \midrule
     1 & Full Model               & \textbf{26.11}  & \textbf{12.93}  & \textbf{0.864}      \\
     \midrule
     2 & w/o Tex. Prior           & 26.01  & 14.35  & 0.863      \\
     \midrule
     3 & w/o Tex. Fitting         & 25.87  & 13.11  & 0.863      \\
     4 & w/o Inversion            & 25.47  & 13.26  & 0.860      \\
     \midrule
     5 & w/o Color Calib.         & 26.05  & 13.09  & 0.862      \\
     6 & w/o Regularization       & 25.76  & 13.56  & 0.857      \\
    \bottomrule
  \end{tabular}
  }
  \caption{Ablation of one-shot reconstruction on InterHand2.6M.}
  \label{tab:ablation-OSR}
\end{table}

\noindent
\textbf{One-shot Reconstruction.}
We show the effectiveness of our strategies for one-shot reconstruction in \cref{tab:ablation-OSR} (on InterHand2.6M as \cref{sec:comparisons}) and \cref{fig:ablation} (in-the-wild).
When there is no texture prior (\#2), a significant drop in LPIPS is observed, indicating that details of the hands are missing (\cref{fig:ablation}).
When omitting texture fitting (\#3), the performance slightly drops, indicating that our inversion strategy can optimize similar hands from the latent space when the training data is adequate.
When not performing inversion (setting the identity features are zero vectors, \#4), the results become worse.
This further shows that inversion can provide better prior knowledge for the one-shot reconstruction.
When ignoring color calibration (\#5), the performance will degenerate even with the data captured under the same camera setups.
When using in-the-wild data for reconstruction~(\cref{fig:ablation}), adopting color calibration is of more significance.
We also demonstrate that overfitting the input image without view regularization leads to poorer quality (\#6).
Finally, we qualitatively show that our self-occlusion modeling with a shadow field can contribute to more realistic hand avatar animation (\cref{fig:ablation}).

\subsection{Robustness towards Diverse Inputs}
To show the robustness of OHTA, we present our quantitative and qualitative performance across diverse inputs.

\noindent
\textbf{Quantitative.}
As depicted in \cref{fig:model-analysis-robustness-input}, OHTA gets similar PSNR and LPIPS for different inputs on InterHand2.6M.

\noindent
\textbf{Qualitative.}
In \cref{fig:model-analysis-in-the-wild}, we illustrate our results on various real-captured images and whole-body images of the MSCOCO dataset \cite{lin2014microsoft,jin2020whole}. 
OHTA performs consistently across different skin tones, hand poses, and viewpoints for real-captured data.
Also, OHTA can reconstruct hand avatars from human images on MSCOCO, which further justifies the robustness of OHTA.
We refer the reader to the supplementary material for additional qualitative results.

\subsection{Application}
\label{sec:application} 
Based on our one-shot framework, we can achieve 1) hand avatar creation with text prompts~\cite{zhang2023adding}, 2) hand avatar editing of the geometry and appearance, and 3) identity-space manipulation including appearance sampling and interpolation. 
The results are shown in \cref{fig:teaser} and \cref{fig:application}.
For more details, please kindly refer to our supplementary materials. 

\noindent
\textbf{Text-to-avatar.} We use ControlNet \cite{zhang2023adding} to generate a hand image with the hand mask and text prompts, then utilize OHTA to reconstruct hand avatars from the hand images.

\noindent
\textbf{Editing.}
The geometry is based on the mesh scaffold.
Thus, we can edit the hand shape parameters $\bm{\beta}$ to edit the geometry.
For appearance editing, we can draw arbitrary content on a target view of the hand avatar and then use OHTA to update the edited content part with its mask.

\noindent
\textbf{Latent space manipulation.}
Our identity latent space is continuous.
Hence, we can sample the identity code from a normal distribution to sample hand avatars.
Furthermore, we can interpolate two target identity codes to obtain smooth appearance interpolations between two identities.

\section{Conclusion}
In this paper, we present the first one-shot implicit hand avatar creation approach, OHTA. 
OHTA consists of two stages. 
The first stage focuses on learning the hand prior on datasets with multiple identities. 
For the creation of one-shot hand avatars, it is only necessary to optimize the second stage with the learned prior, which includes texture inversion and texture fitting.
OHTA is robust for diverse input images and is capable of solving various downstream tasks with consistent animations. 
These tasks include hand avatar creation with text prompts, hand geometry and appearance editing, and manipulation of identity latent space.

\noindent
\textbf{Limitations and Future Works.}
OHTA exhibits suboptimal performance for input images that feature 1) notably uneven lighting, and 2) highly inaccurate pose estimation.
Thus, further designing the approach to be more robust for those situations is worthy of ongoing exploration.
\maketitlesupplementary

\renewcommand{\thesection}{\Alph{section}}  
\renewcommand{\thetable}{\Alph{table}}  
\renewcommand{\thefigure}{\Alph{figure}}

\setcounter{section}{0}
\setcounter{figure}{0}
\setcounter{table}{0}

\noindent
In the supplemental material, we provide:
\begin{itemize}
\item[\S]\ref{suppsec:details}: Implementation Details.
\item[\S]\ref{suppsec:exp}: More Experiments and Results.
\item[\S]\ref{suppsec:discussions}: Discussions.
\item[\S]\ref{suppsec:vis}: More Qualitative Results.
\end{itemize}

\section{Implementation Details}
\label{suppsec:details}

\subsection{Spatial Interpolation}
We uniformly sample $N^{p}_{k}$ anchor points $\mathbf{P}_{k}$ on the input mesh and represent them with barycentric coordinates for each resolution, where $k \in \{1, ..., K\}$ denotes the resolution level. 
The sampled barycentric coordinates are fixed after the sampling, which means the sampling for each resolution is only conducted once for obtaining the fixed point encodings $\mathbf{E}_{k}$.
That means that $\mathbf{P}_{k}$ are 3D points representing the location of $\mathbf{E}_{k}$.

\begin{figure}[h!]
    \centering
    \includegraphics[width=\columnwidth]{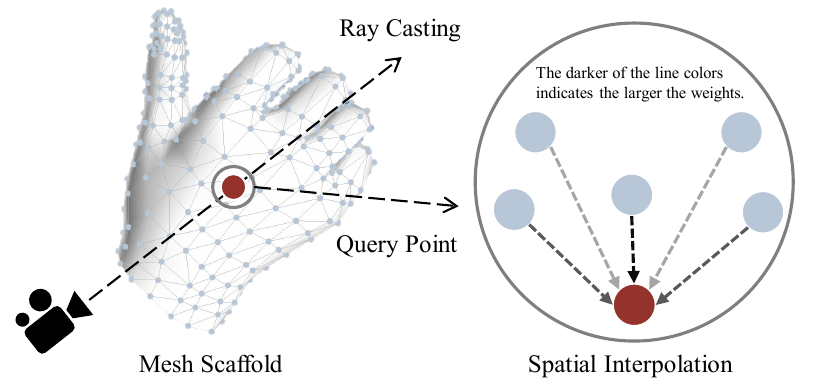}
    \caption{Illustration of spatial interpolation for a level of resolution. }
    \label{fig:supp-interpolation}
\end{figure}

For query points $\mathbf{q}$, we conduct spatial interpolation to acquire the queried encoding of this resolution:
\begin{equation}
    \mathbf{Q}_{k}=\text{interp}(\mathbf{q}, \mathbf{E}_{k}).
\end{equation}
\cref{fig:supp-interpolation} depicts the spatial interpolation process.
Specifically, we first extract the encodings of $N^{n}$ neighbor points $\mathcal{K}(\mathbf{E}_{k}) \in \mathbb{R}^{N^{n} \times N^{q} \times N^c}$ of $\mathbf{P}_{k}$, where $\mathcal{K}$ denotes \textit{k}-nearest neighbors. 
Then, we perform a weighted average with the inverse Euclidean distances of those points to the query point as the weights (values indicated by the \textbf{line colors} in \cref{fig:supp-interpolation}) to acquire the queried encoding of this resolution $\mathbf{Q}_{k} \in \mathbb{R}^{N^{q} \times N^{c}}$ (\ie the features of the \textcolor{red}{red point} in \cref{fig:supp-interpolation}).

\begin{figure}[h!]
    \centering
    \includegraphics[width=\columnwidth]{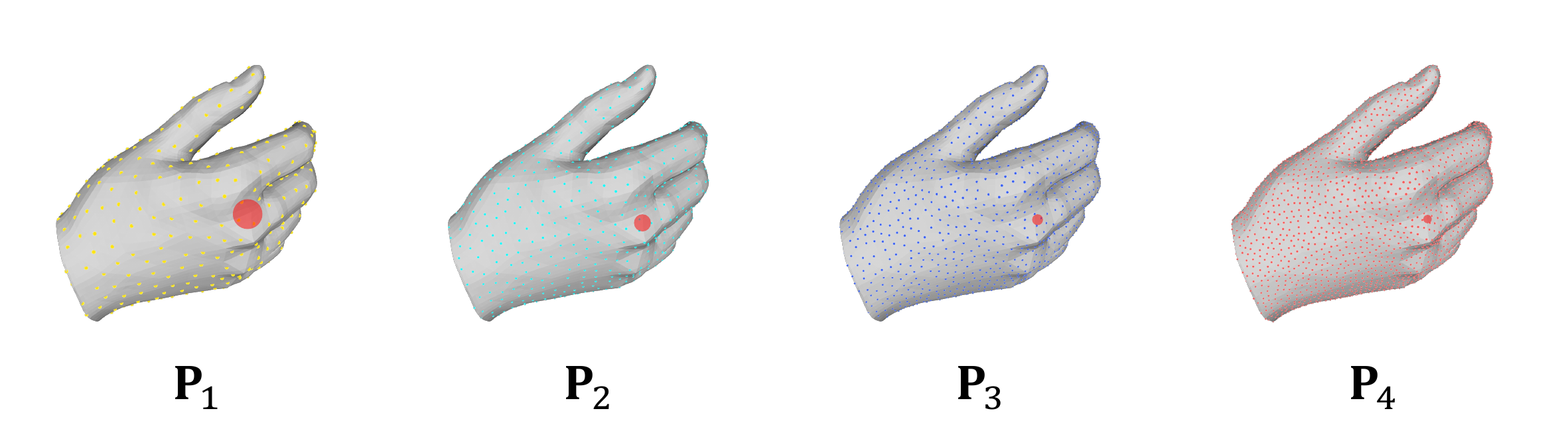}
    \caption{Illustration of $\mathbf{P}_{k}$ in different level of resolutions. The light red area in the figure indicates the region used for spatial interpolation for a query point.}
    \label{fig:supp-pk}
    \vspace{-0.5cm}
\end{figure}

Furthermore, to illustrate the impact of different resolutions on spatial interpolation, we show the anchor points $\mathbf{P}_{k}$ used in multi-resolution fields at different resolutions in~\cref{fig:supp-pk}. The light red area in the figure indicates the region used for interpolation to obtain $\mathbf{Q}_{k}$.
For more discussion on multi-resolution, please refer to \cref{sec:resolution}.

\subsection{Network Structure}
\noindent
\textbf{Architecture.}
For shape fitting, $\mathcal{M}_{shape}$ consists of 4 layers with a hidden dimension of 128.
For multi-resolution fields, $\mathcal{M}_{k}$ contains 4 layers with a hidden dimension of 256, and $\mathcal{M}_{fuse}$ contains 3 layers with a hidden size of 64.

\noindent
\textbf{Model Size.}
The trainable parameters of the model are 4.46M in total.

\subsection{Used Data}
\textbf{InterHand2.6M~\cite{moon2020interhand2}.}
For hand prior learning, we utilize `train/Capture0', `train/Capture1', `train/Capture2', `train/Capture3', `train/Capture5', `train/Capture6', `train/Capture7', `train/Capture8', `train/Capture9', `train/Capture10', `train/Capture11', `train/Capture12', `train/Capture13', `train/Capture14', `train/Capture15', `train/Capture16', `train/Capture20', `train/Capture22', `train/Capture23', `train/Capture24', and `train/Capture25' for training, with one capture for an identity.
Moreover, we use `0000\_neutral\_relaxed', `0009\_thumbtucknormal', `0019\_alligator\_closed', `0029\_indextip', `0039\_fingerspreadrigid', `0048\_index\_point', and `0058\_middlefinger' for evaluation and other poses for training.
For evaluation of one-shot avatars, we employ `test/Capture0/ROM03\_RT\_No\_Occlusion', following HandAvatar \cite{chen2023hand}.
For each frame, we crop the hand region with annotated detection boxes as the ground truth, which is consistent with HandAvatar. 
Specifically, the box is firstly regulated as a square box with 1.3 times expansion. Then, the hand region is cropped and resized to $256\times 256$ resolution.
Unless otherwise stated, we all adopt the same cropping approach for experiments of all the used data.

\noindent
\textbf{HanCo~\cite{zimmermann2021contrastive}.}
For quantitative experiments on the HanCo dataset, we utilize sequence `0191' with the camera above the hands (cam3, cam5, cam6, and cam7). 
We do not adopt cameras below the hands because the capturing environment exhibits uneven lighting conditions and inconsistent color calibration, which causes significantly inconsistent appearances of the hands for the images captured below.
We utilize the MANO annotations and the provided hand masks of this dataset for one-shot reconstruction.

\noindent
\textbf{MSCOCO~\cite{lin2014microsoft}.}
To test OHTA's performance for the challenging in-the-wild images, we take the whole-body version of MSCOCO \cite{jin2020whole} for experiments.
We utilize the pose estimation results provided by InterWild \cite{moon2023bringing} and generate the masks using SAM \cite{kirillov2023segment}.

\noindent
\textbf{OneHand10K~\cite{wang2018mask}.}
In addition to MSCOCO, we also provide in-the-wild results on the OneHand10K dataset, which is one of the largest monocular hand pose estimation datasets.
Since OneHand10K does not provide labels for 3D pose estimation of the hand, we utilize DIR \cite{ren2023decoupled} for pose estimation and use the corresponding pose estimation results to generate hand masks.

\noindent
\textbf{Real-captured Data.}
We also capture images of hands with different identities and poses, ensuring a large demographic diversity among the subjects to show the robustness of OHTA.
The hand pose estimation results and masks for those images are obtained by DIR~\cite{ren2023decoupled}.

\subsection{Hand Prior Learning}

The identity codes $\mathbf{z} \in \mathbb{R}^{21\times33}$ are learnable parameters initialized from a truncated normal distribution with standard deviation $\sigma=0.02$, where 21 denotes that we utilize 21 subjects of InterHand2.6M \cite{moon2020interhand2} for training and 33 denotes the dimension of an identity's code.

We follow the implementation of HandAvatar \cite{chen2023hand} to pre-train PairOF.
If not stated otherwise, we adopt Adam optimizer \cite{kingma2014adam} for optimization.
The learning rate begins at $5e^{-4}$ and decreases with exponential decay. 
The training process has 300K steps with a batch size of 32.

For end-to-end prior learning, we use a patch strategy for training with a patch size $32 \times 32$, following \cite{weng2022humannerf, chen2023hand}.
The learning rate begins at $5e^{-4}$ and decreases with exponential decay. 
The complete prior learning process takes 300K steps with a batch size of 2.
Since the masks from MANO are not well aligned with the hands in the image, using those masks for personalized shape fitting will result in underperforming learning of texture prior.
Therefore, we adopt SAM \cite{kirillov2023segment} for mask prediction since it can obtain hand contour-aligned results.
Specifically, we utilize the ViT-H version with 2D joint positions as the point prompts for predictions.

\subsection{One-shot Reconstruction}
\label{sec:one-shot reconstruction}
\textbf{Texture Inversion.}
For inversion, the input and rendered image resolutions are $256 \times 256$. 
We initialize the identity code as zero vectors for optimization.
The complete inversion process takes 50 steps with the learning rate of $1e^{-2}$.
Since the fingernails are not related to color calibration and might dominate the inversion, we mask the fingernails for optimization.
Specifically, we utilize the classified occupancy values of PairOF \cite{chen2023hand} to derive the masks of the fingernails.

\noindent
\textbf{Texture Fitting.}
For fitting, we use a patch strategy with a patch size $128 \times 128$ for optimization.
The complete inversion process takes 100 steps with the learning rate of $1e^{-3}$.
The $N^{r}$ reference views for view regularization are generated by uniformly rotating the hands around the axis of the MCP joint of the middle finger and wrist with each rotation angle of $\frac{2\pi}{(N^{r} + 1)}$. 
The hand pose of the reference views are set to flatting hand (canonical pose of MANO).
Considering the fingernails may vary a lot from the inversion result to the fitting target, we make use of an additional loss for the fingernails using the fingernails mask derived from the PairOF.

\begin{figure}
    \centering
    \includegraphics[width=0.47\textwidth]{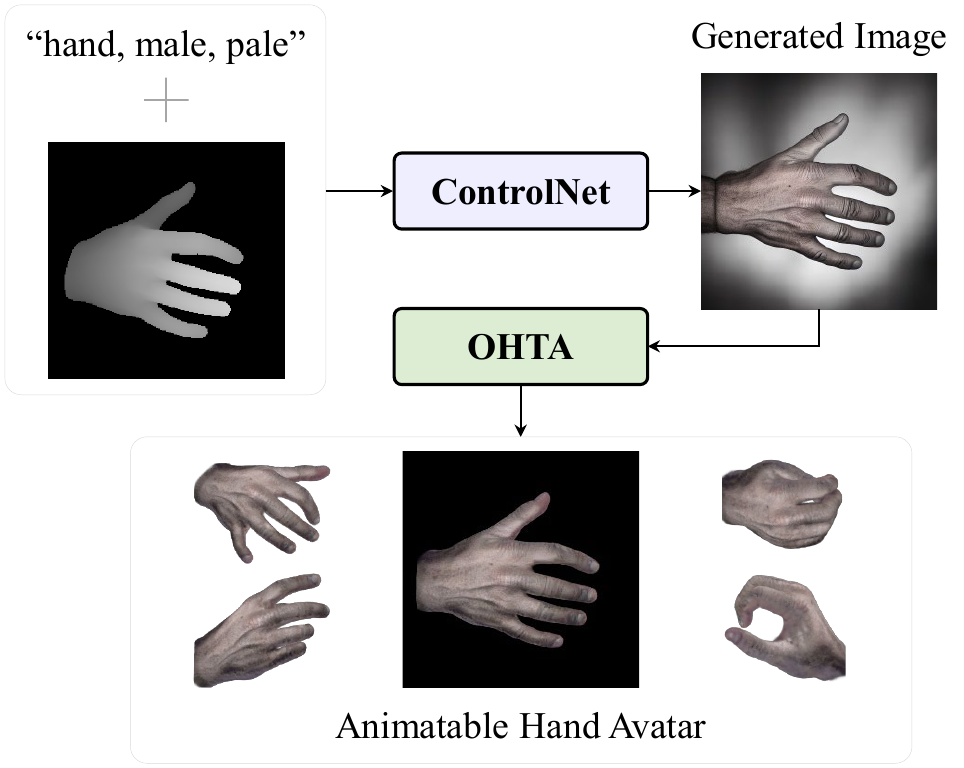}
    \caption{Hand avatar creation with text prompts. }
    \label{fig:supp-text2avatar}
\end{figure}

\subsection{Application}
\noindent
\textbf{Text-to-avatar.} 
The complete pipeline for text-to-avatar is illustrated in \cref{fig:supp-text2avatar}.
We use ControlNet 1.1 \cite{zhang2023adding} with depth maps as inputs for hand image generation.
The generation is guided by the depth map of the back of the hand and input text prompts, using the default parameters of the model.
After generation, we utilize OHTA to reconstruct hand avatars from the hand images.
The optimizing process follows the same procedure of one-shot reconstruction (\cref{sec:one-shot reconstruction}) except that using 200 steps for texture fitting to better capture complex details of the generated hands.

\noindent
\textbf{Editing.}
On account of our geometry based on the mesh scaffold, we can edit the geometry of avatars by changing the guided mesh scaffold.
Since we utilize the MANO for guidance, we can modify the shape parameters $\bm{\beta}$ of the MANO to edit the geometry.
For appearance editing, we can first render a target view of the hand avatar and draw desired content on the rendered image.
Then, we can utilize OHTA for one-shot reconstruction.
This process does not require texture inversion and only makes use of 100 steps' texture fitting with the edited contents and the corresponding mask to update the edited parts.

\noindent
\textbf{Latent space manipulation.}
With multiple identities for training, we obtain a continuous latent space.
Therefore, we can conduct latent space manipulation, including latent space sampling and interpolation.
For sampling, we randomly sample an identity code $\mathbf{z^{\prime}} \in \mathbb{R}^{33}$ from a normal distribution.
Then, we can use $\mathbf{z^{\prime}}$ to obtain the sampled avatar.
For interpolation, we can take two identity codes $\mathbf{z_{1}}$ and $\mathbf{z_{2}}$ for combinations: $\mathbf{z^{\prime\prime}} = t\mathbf{z_{1}} + (1 - t)\mathbf{z_{2}}$, where $t \in [0, 1]$.
With the interpolated identity code $\mathbf{z^{\prime\prime}}$, we can obtain the hand avatar with the interpolated appearance.

\section{More Experiments}
\label{suppsec:exp}

\begin{figure}[h!]
    \centering
    \includegraphics[width=0.47\textwidth]{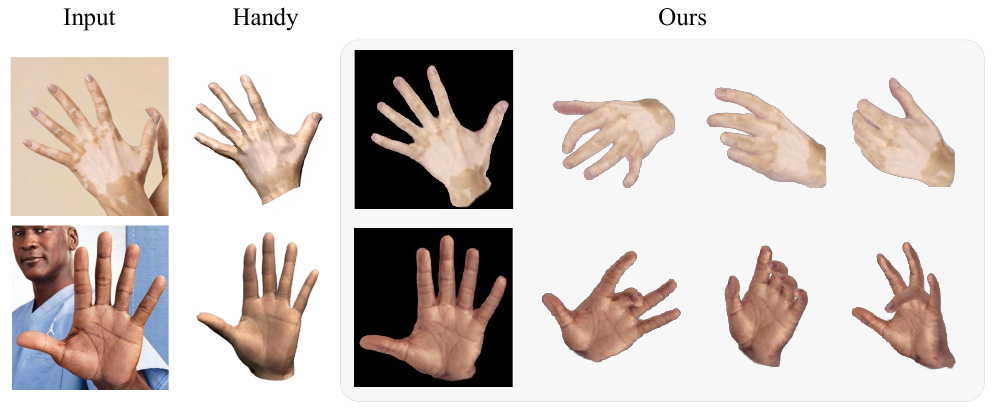}
    \caption{Comparison with Handy on the results reported in the original paper.}
    \label{fig:supp-comparison-handy}
\end{figure}

\textbf{Comparison with Handy.}
As shown in \cref{fig:supp-comparison-handy}, we compare our results with the in-the-wild results presented in the Handy~\cite{potamias2023handy} paper. The experimental results demonstrate that OHTA is capable of effectively modeling variations in skin tone and the details of the hand.

\begin{figure}[h!]
    \centering
    \includegraphics[width=0.47\textwidth]{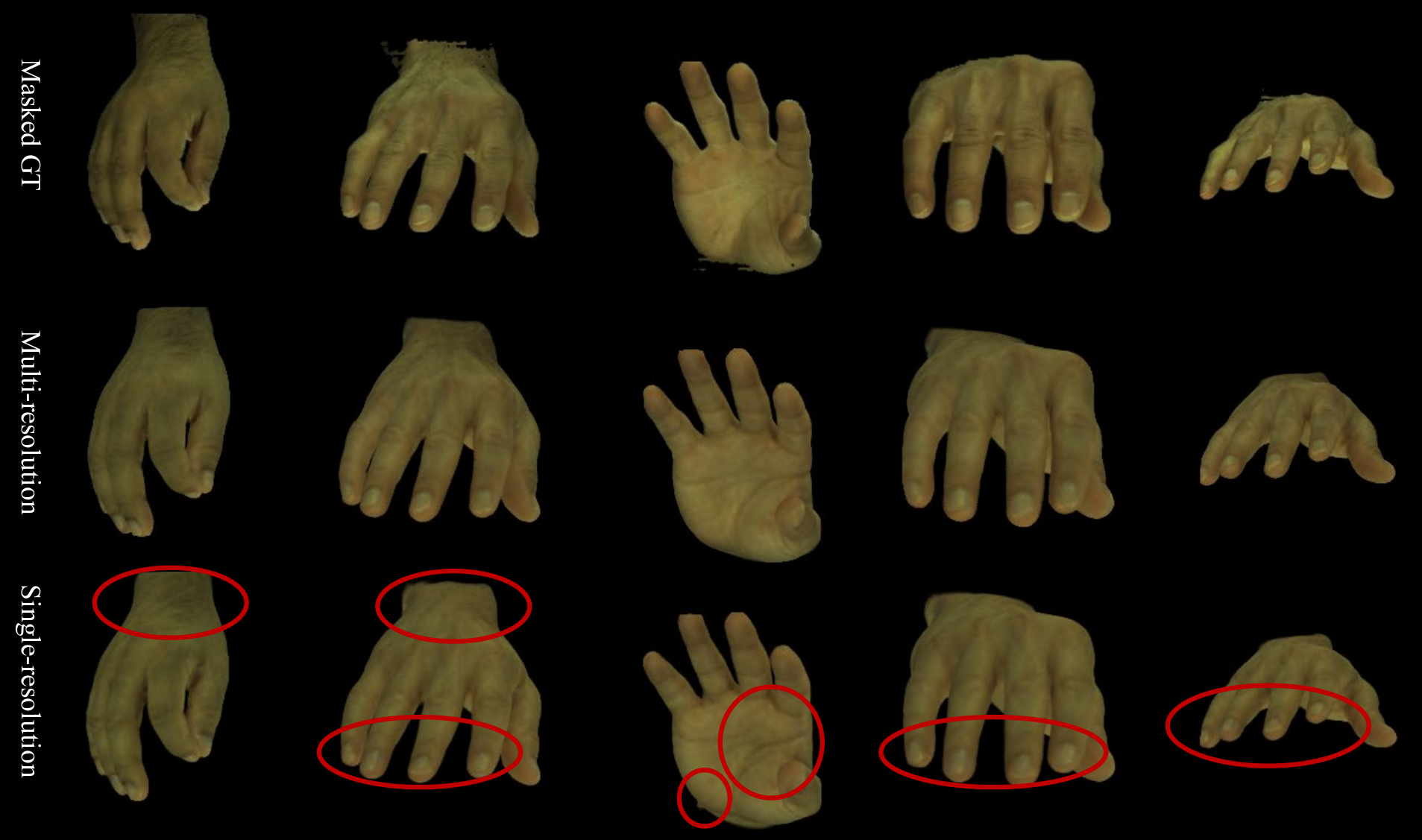}
    \caption{Comparison between using multi-resolution and single-resolution. The \textcolor{red}{red ellipse} indicates those not well-captured details by single-resolution.}
    \label{fig:supp-resolution}
\end{figure}

\noindent
\textbf{Different Resolution.}\label{sec:resolution}
Using high-resolution encodings with dense points is able to model details of the texture, while low-resolution encodings with sparse points are capable of modeling the long-range dependencies of the texture for a more consistent appearance.
Therefore, we conduct experiments to tune the best resolution combinations.
\cref{tab:res. level} shows the performance of Hand Prior Network (HPNet) with different resolutions for the albedo prior learning.
We do not take resolutions larger than 4096 points since it will introduce too much computation.
The results show that using 4 resolutions performs best.
More resolutions (\eg, 8 resolutions) degenerating the performance may be because 1) resolutions of too sparse points (\eg, $\{32 \times 2^{k-1}\}_{k=1}^{4}$) are not able to encode detailed information is beneficial for the performance and 2) too many resolutions may lead to optimization difficulty.
Therefore, we adopt 4 resolutions for our HPNet.
The qualitative comparisons between using multi-resolution and single-resolution are shown in \cref{fig:supp-resolution}. 
The results are tested with unseen poses. 
From those results, we can see that using multi-resolution is able to model more details of the hands and makes the learned overall appearance more consistent with the ground-truth.

\begin{table}[h!]
  \centering    
  \begin{tabular}{cccccccc}
    \toprule 
     Used Resolution      & PSNR & LPIPS & SSIM   \\
    \midrule
     $\{4096\}$                                    & 27.11 & 12.96 & 0.890     \\
     $\{512, 1024\}$                               & 27.18 & 12.91 & 0.890     \\
     $\{512, 4096\}$                               & 27.46 & 12.80 & 0.895     \\
     $\{512 \times 2^{k-1}\}_{k=1}^{4}$            & \textbf{27.64} & \textbf{12.23} & \textbf{0.896}     \\
     $\{32 \times 2^{k-1}\}_{k=1}^{8}$             & 27.32 & 12.28 & 0.894     \\
    \bottomrule
  \end{tabular}
  \caption{Comparison of using different resolution combinations for the albedo field.}
  \label{tab:res. level}
\end{table}

\begin{figure}[h!]
    \centering
    \includegraphics[width=0.47\textwidth]{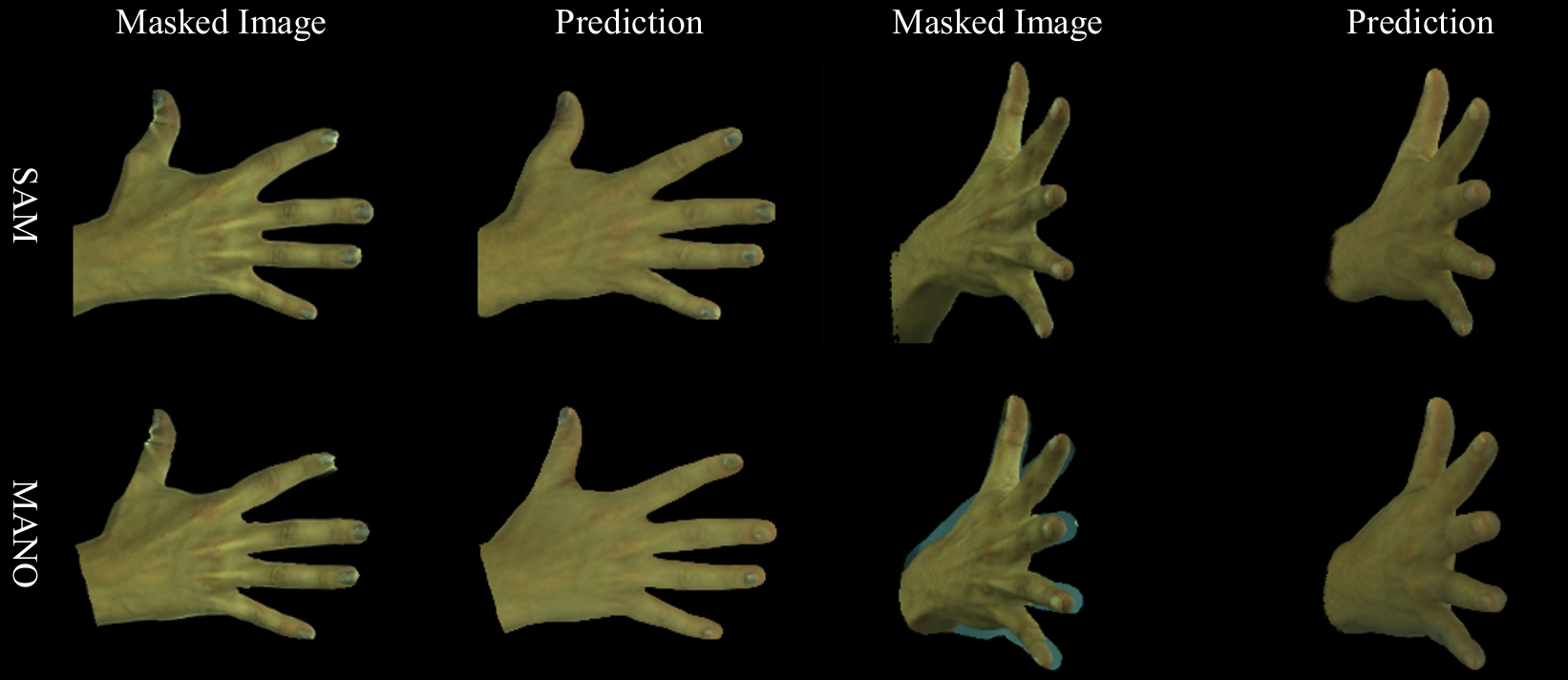}
    \caption{Comparison between using different masks. }
    \label{fig:supp-mask}
\end{figure}

\noindent
\textbf{Different Mask.}
As shown in \cref{fig:supp-mask}, using masks better aligned with the images is beneficial for personalized shape fitting during hand prior learning stage.
When the hand geometry is refined to be better aligned with the input images using the masks from SAM, the learned texture prior can better capture the details of the hand to improve the fidelity.

\section{Discussion}
\label{suppsec:discussions}
\subsection{Mesh-guided Representation}
We adopt mesh-guided representation for several reasons.
For the geometry, the implicit occupancy field learned with the mesh information is more robust for novel identities, as proved in \cite{mihajlovic2022coap}. 
For the texture, the implicit texture field learned on the mesh scaffolds can be transferred to other hand shapes, which is shown by our geometry editing capability.
This nature is important for the one-shot reconstruction since the hands have a large variety in shape. 
Other texture representations (e.g., volume-based) \cite{mundra2023livehand,guo2023handnerf} have poor performance for novel hand shapes since they learn the neural hand representations in canonical space that have the fixed hand size.
Moreover, this representation is more robust for one-shot fitting since it utilizes several neighboring anchors' features for prediction to obtain relatively transition-smooth results.

\begin{figure}[t!]
    \centering
    \includegraphics[width=0.47\textwidth]{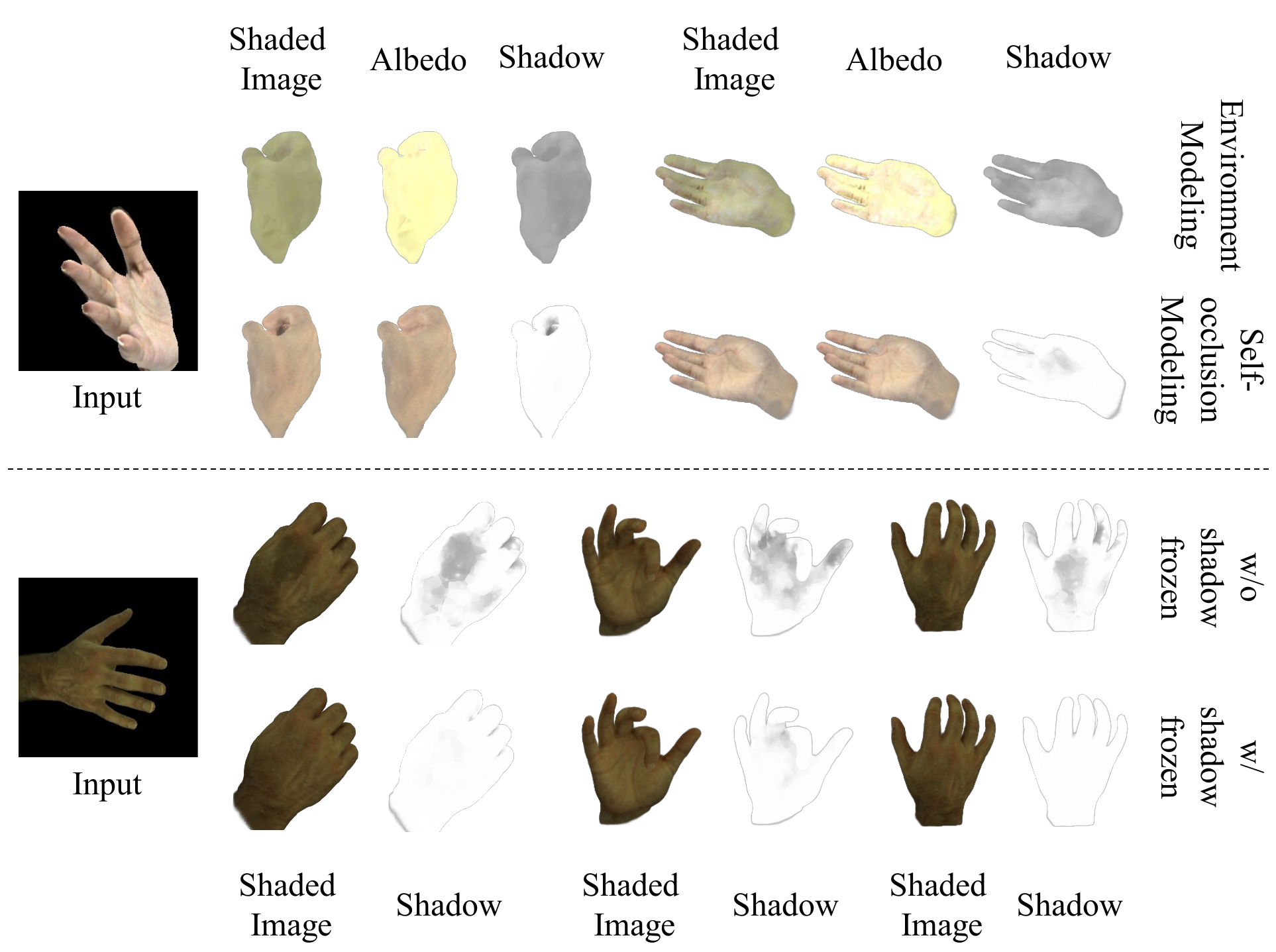}
    \caption{Comparison between using different shadow approaches for one-shot reconstruction. The above shows the comparisons between environment modeling and our self-occlusion modeling. The below illustrates the differences between whether to freeze the shadow field for one-shot reconstruction.}
    \label{fig:supp-shadow}
\end{figure}

\subsection{Shadow Modeling}
\label{suppsec:discussion-shadow}
HPNet separates texture modeling into albedo and shadow modeling, following a foundational principle similar to that of HandAvatar~\cite{chen2023hand}.
However, our design diverges significantly from HandAvatar, particularly in its motivation and network structure.

For motivation, we aim to learn transferable shadow prior knowledge for one-shot hand avatar creations, while HandAvatar learns the illumination fields for a specific hand that are not transferable.
To this end, our shadow branch is designed to be identity-shared and view-independent, which can learn transferrable general self-occlusion effects that are shared across different identities. 
The reasons for this design are twofold.
On one hand, it is non-trivial to learn the environmental lighting conditions based on a single image. 
Thus, we should rely on shadow prior learned from training data to generate plausible shadows for one-shot hand avatar creations.
On the other hand, the identity-shared and view-independent shadow prior can be transferred to one-shot creation, while the identity-specific and view-conditioned prior learned by environmental lighting modeling cannot.

The specific shadow network structure of HPNet includes the identity-shared design, hand finger pose without global rotation as the pose condition, and the same multi-resolution fields as our albedo field.
In comparison, HandAvatar has a dedicated design for an illumination field with positional encodings, directed soft occupancy, and a pose with global rotation as input, enabling it to capture the environmental lighting conditions for different hand poses.

As shown in \cref{fig:supp-shadow}, our learned shadow prior can be effectively transferred to the novel identity to predict the identity-shared self-occlusion effects when there is no adequate knowledge of the real environmental lighting condition.
In contrast, using identity-specific view-conditioned environment modeling like HandAvatar \cite{chen2023hand} fails in reconstructing the novel hand avatar with plausible shadows.

\begin{figure}[t!]
    \centering
    \includegraphics[width=0.47\textwidth]{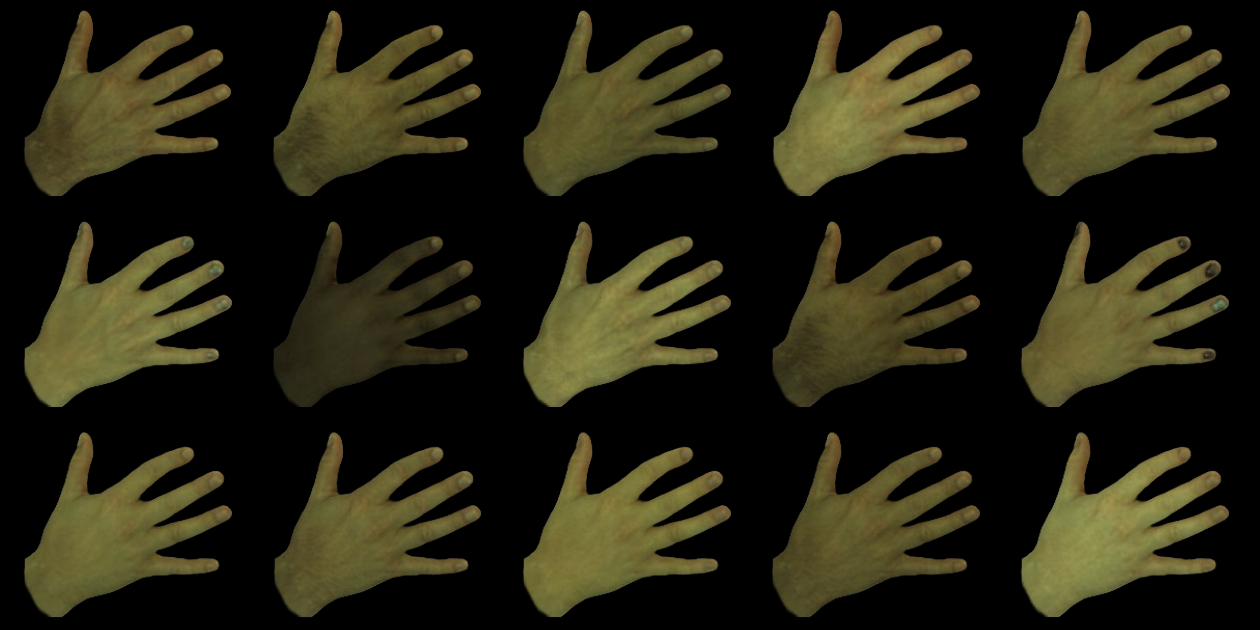}
    \caption{Part of the learned identities in the prior learning stage. }
    \label{fig:supp-identity}
\end{figure}

\subsection{Difference between hand and other body parts}
While recent studies have focused on one-shot human body or head avatars, they are not suitable for hand modeling due to inherent task differences.
Created hand avatars aim to produce highly consistent animations for different poses and viewpoints, which is not similar to head avatars that concentrate less on the performance around the backside of the heads and are even not concerned about the animation like Preface \cite{buhler2023preface}.
Human avatars also require animation considerations, yet existing methodologies are inadequate for hand animations.
In the case of explicit approaches employing generative models, such as DINAR \cite{svitov2023dinar}, our experiments detailed in the main text for Handy \cite{potamias2023handy} show that these methods fall short in generating high-fidelity hand avatars.
Approaches employing diverse model priors like ELICIT \cite{huang2022elicit} mainly utilize the symmetry of human body appearance for texture modeling.
This kind of method is not appropriate for hands since modeling the hand appearance is mainly about modeling the details of the hands rather than only the hand skin tone. 
Those priors are not enough to model invisible hand details.
Another type of generalizable NeRF approach, like SHERF \cite{hu2023sherf}, may also be inapplicable for one-shot hand avatar creation due to its inability to model high-quality unobserved parts of hands.

\subsection{Discussion about OHTA and PhoneScan}
PhoneScan~\cite{cao2022authentic} also proposes using prior models coupled with fine-tuning to achieve few-shot reconstructions.
We provide a discussion about the similarities and differences between OHTA and PhoneScan.

\noindent
\textbf{Similarity}: Both OHTA and PhoneScan 1) embed the priors to the model by training on large-scale data, and 2) include a fine-tuning process for adjusting the pre-trained model to fit the target images.

\noindent
\textbf{Difference}: 
OHTA differs from PhoneScan in various aspects.
\noindent
1) \emph{motivation}: OHTA aims at addressing hand avatar creation from a \textbf{single} RGB image, while PhoneScan focuses on solving head avatar creation from RGB-D videos. Even with methods shared similarities, this essential difference leads to different design priorities (described below); 
\noindent
2) \emph{optimization pipeline}: OHTA reconstructs with inversion and fitting, which highly relies on inversion from identity space for regularization of the fitting.
As described in~\cite{cao2022authentic}, PhoneScan has no identity space, predicts person-specific information (\ie bias maps) with the pre-trained model, and fine-tunes the model with inputs.
As mentioned in the ablation of~\cite{cao2022authentic}, there exist artifacts for novel views when only fine-tuning with frontal images. 
In contrast, OHTA exhibits robust performance for one-shot creation.

\noindent
3) \emph{prior representation}: OHTA exploits identity codes and MLPs for geometry, albedo, and shadow prior learning, while PhoneScan embeds identity and expression prior knowledge in pre-trained CNNs.

\subsection{Correlation between input and identity codes}
\begin{figure}[h!]
    \centering
    \includegraphics[width=\columnwidth]{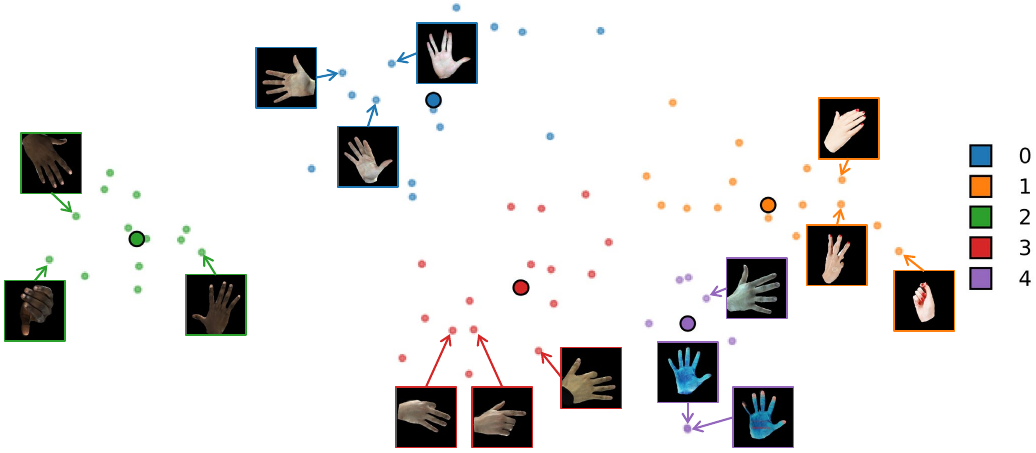}
    \caption{t-SNE visualization of optimized identity codes from diverse images. Labels of clustering (colored points) are shown for visualization.}
    \label{fig:tsne}
\end{figure}

In \cref{fig:tsne}, we show the identity codes distribution with different inputs, using t-SNE to embed the identity codes into 2D space.
We found that similar inputs have closer distances in the identity space.
For one-shot creation, identity codes 1) only relate to the hand albedo and 2) do not affect geometry, based on our hand representation design.
Fig.~\textcolor{red}{1} in the main text demonstrates that our identity space is continuous, allowing for interpolation between two hand identities while maintaining the hand geometry.
\cref{fig:supp-shape-editing} in supp. shows OHTA's capability for shape editing while preserving the identity.

\section{Qualitative Result}
\label{suppsec:vis}

\subsection{Prior Learning Result}
\cref{fig:supp-identity} presents some of the learned identities in the prior learning stage. 
The images are rendered with different identity codes and the same hand pose.
Those results demonstrate that our prior learning is able to capture the details of different identities, which facilitates the one-shot reconstruction.

\subsection{One-shot Result}
\begin{figure*}[t!]
    \centering
    \includegraphics[width=0.99\textwidth]{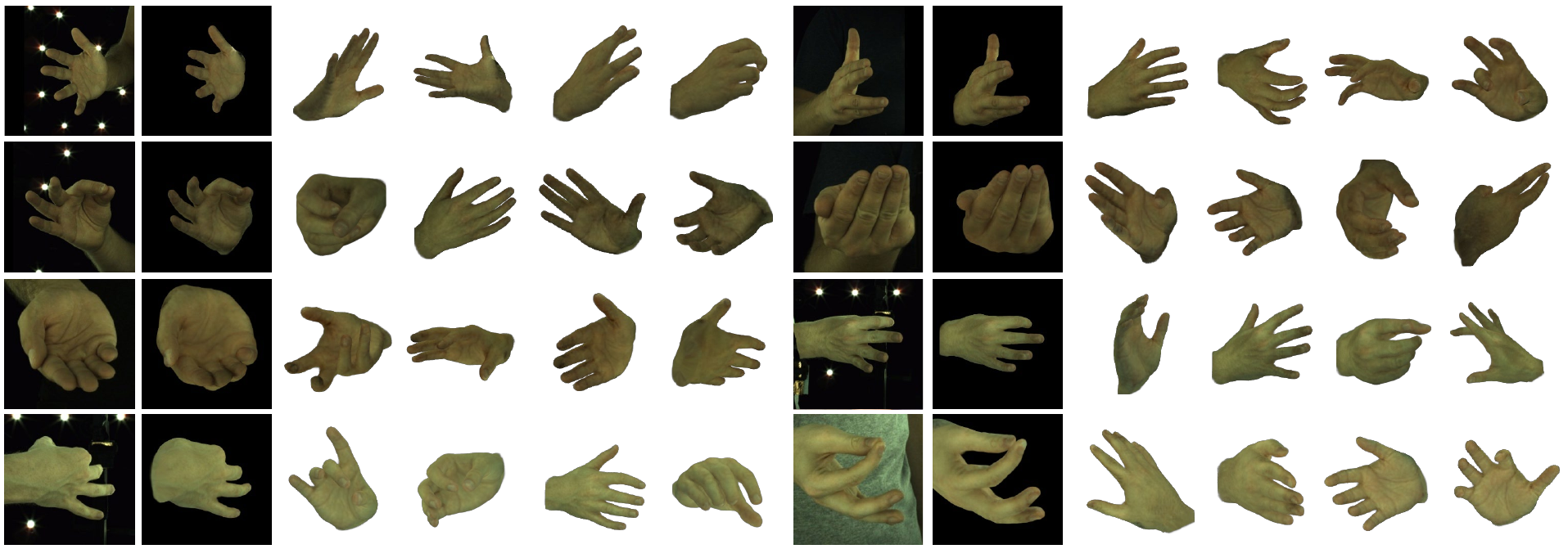}
    \caption{\textbf{Qualitative results on InterHand2.6M~\cite{moon2020interhand2}.} For each example, we show from left to right: (a) the target image, (b) the fitted avatar rendered to the input view, and (c) the results of the hand avatar rendered in novel poses.}
    \label{fig:supp-interhand}
\end{figure*}

\begin{figure*}
    \centering
    \includegraphics[width=\textwidth]{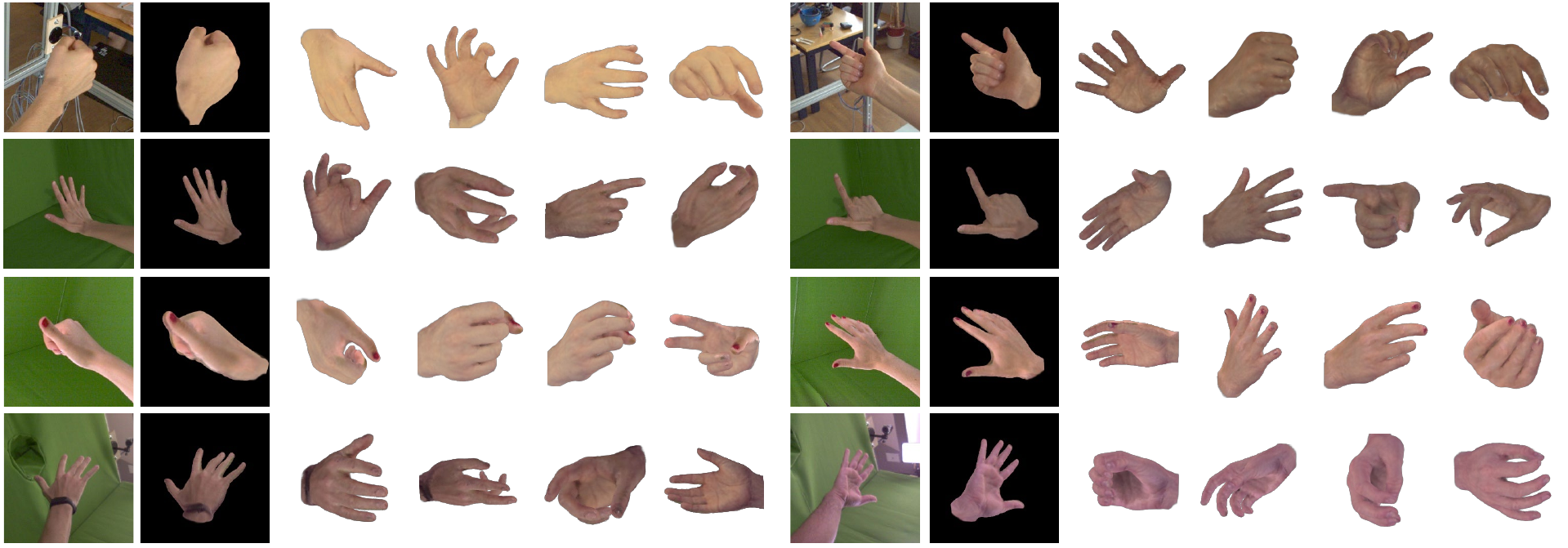}
    \caption{\textbf{Qualitative results on HanCo~\cite{zimmermann2021contrastive}.} For each example, we show from left to right: (a) the target image, (b) the fitted avatar rendered to the input view, and (c) the results of the hand avatar rendered in novel poses.}
    \label{fig:supp-hanco}
\end{figure*}

\begin{figure*}
    \centering
    \includegraphics[width=\textwidth]{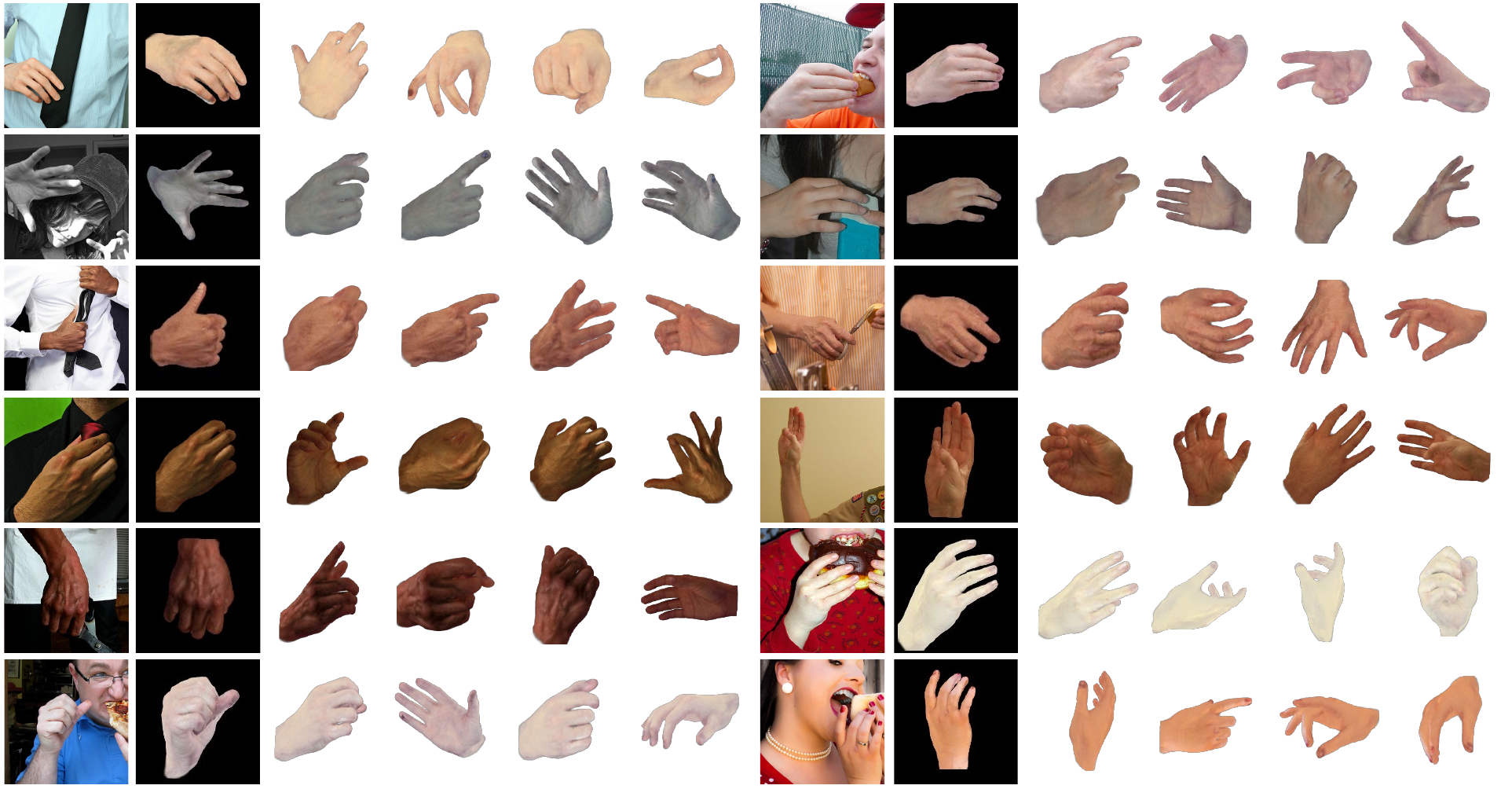}
    \caption{\textbf{In-the-wild results on MSCOCO~\cite{lin2014microsoft}.} For each example, we show from left to right: (a) the target image, (b) the fitted avatar rendered to the input view, and (c) the results of the hand avatar rendered in novel poses.}
    \label{fig:supp-coco}
\end{figure*}

\begin{figure*}
    \centering
    \includegraphics[width=\textwidth]{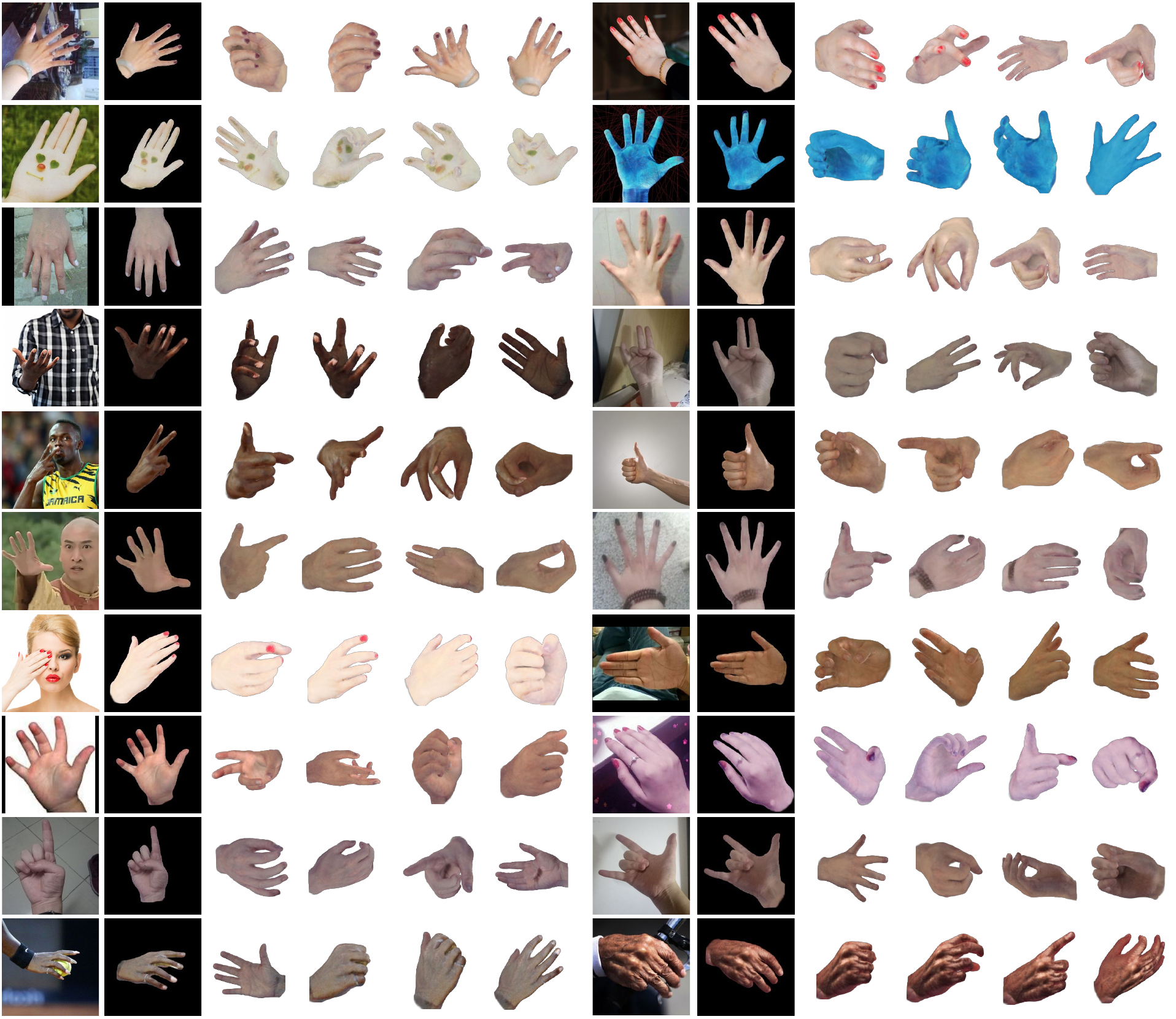}
    \caption{\textbf{In-the-wild results on OneHand10K~\cite{wang2018mask}.} For each example, we show from left to right: (a) the target image, (b) the fitted avatar rendered to the input view, and (c) the results of the hand avatar rendered in novel poses.}
    \label{fig:supp-onehand10k}
\end{figure*}

\begin{figure*}
    \centering
    \includegraphics[width=1.00\textwidth]{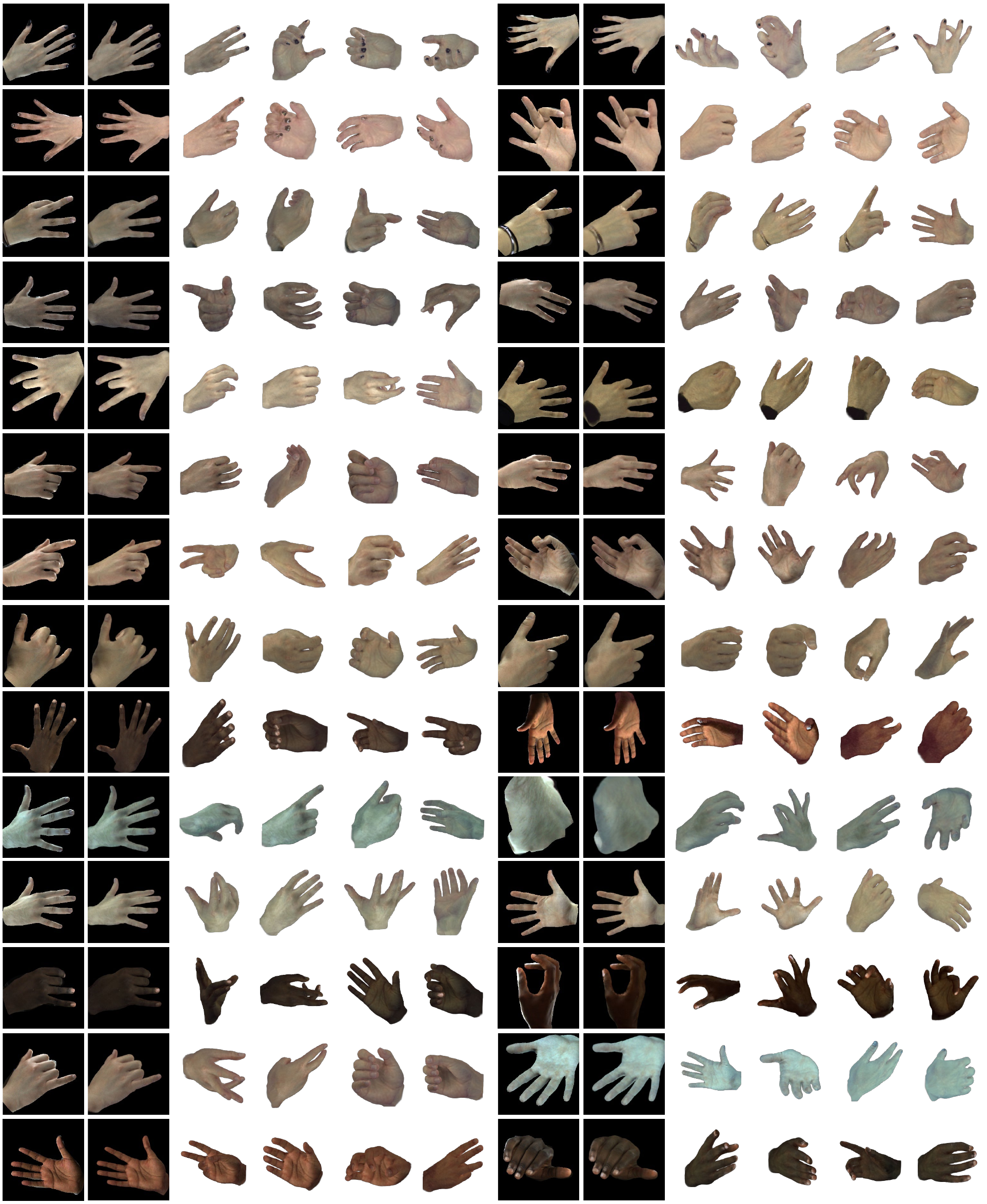}
    \caption{\textbf{In-the-wild results on the real-captured data.} For each example, we show from left to right: (a) the target image, (b) the fitted avatar rendered to the input view, and (c) the results of the hand avatar rendered in novel poses.}
    \label{fig:supp-real-captured}
\end{figure*}

To better show the robustness of OHTA, we show quantities of reconstructed hand avatars from OHTA on InterHand2.6M~\cite{moon2020interhand2}, HanCo~\cite{zimmermann2021contrastive}, MSCOCO~\cite{lin2014microsoft,jin2020whole}, OneHand10K~\cite{wang2018mask}, and real-captured data.
The input images of InterHand2.6M are all from the testing set that has no identity and sample overlap with the prior learning stage.
As shown in~\cref{fig:supp-interhand}, our performance on InterHand2.6M is consistent across diverse hand poses and identities.
We also take lots of samples varied a lot from the hands for prior learning (InterHand2.6M) to further demonstrate the one-shot capability of OHTA.
\cref{fig:supp-hanco} shows more visual results of OHTA on HanCo.
OHTA is quite robust for different poses and viewpoints of the HanCo dataset.
We present more results of OHTA on the MSCOCO dataset in \cref{fig:supp-coco}.
To better validate the robustness of OHTA, we also take lots of samples of the OneHand10K dataset for experiments.
The results are shown in \cref{fig:supp-onehand10k}.
Reconstructing hand avatars for MSCOCO and OneHand10K is challenging because 1) the pose estimations of the hands exhibit apparent deviations from the hands in the images and 2) the hands are with low resolutions.
Despite facing these challenges, OHTA is still robust enough to create hand avatars with consistent animations.
\cref{fig:supp-real-captured} presents more hand avatars reconstructed by OHTA for the real-captured images.

\subsection{Application Result}
\begin{figure*}
    \centering
    \includegraphics[width=1.00\textwidth]{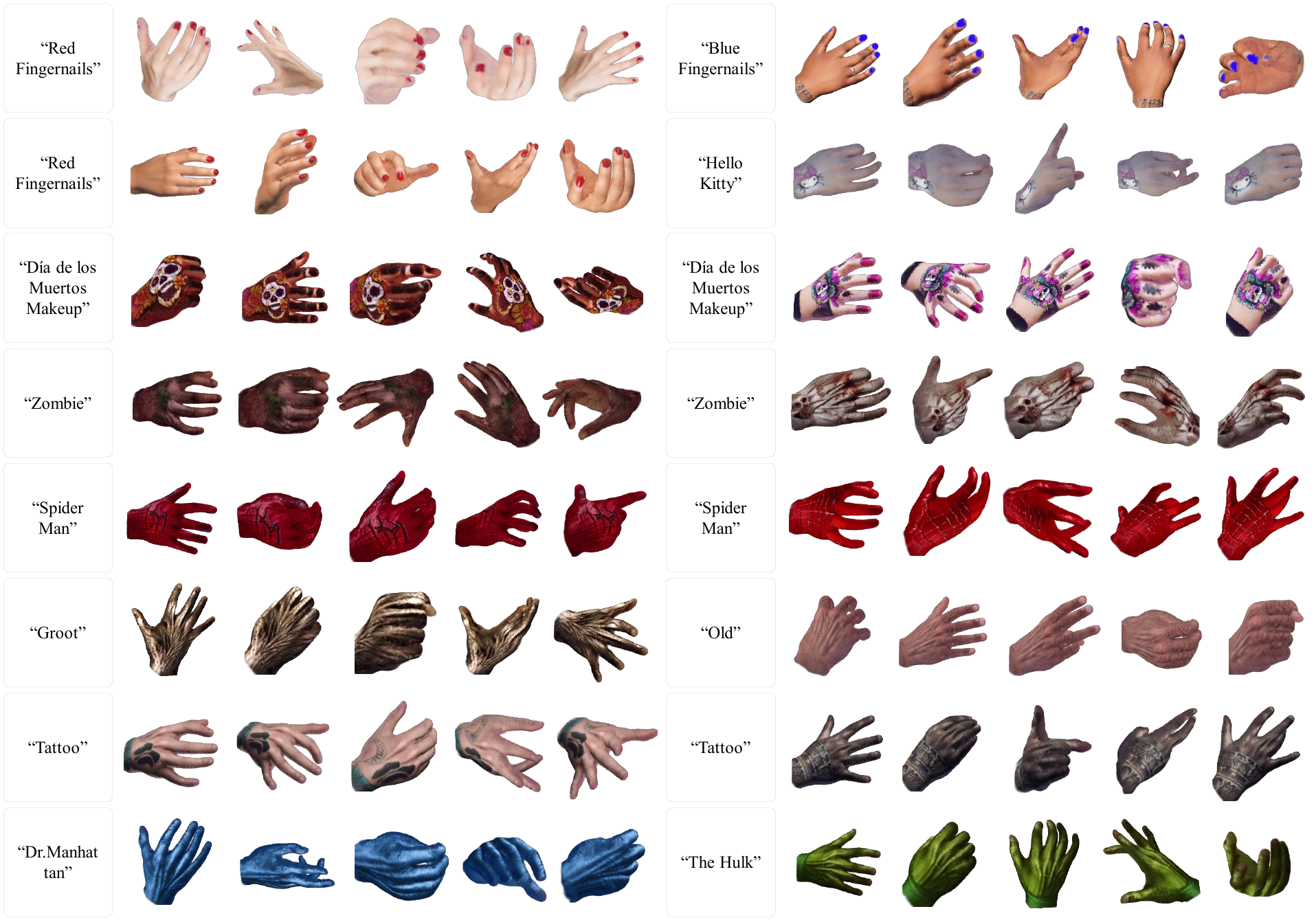}
    \caption{Text-to-avatar results. }
    \label{fig:supp-t2a-result}
\end{figure*}

\begin{figure*}
    \centering
    \vspace{-0.3cm}
    \includegraphics[width=1.00\textwidth]{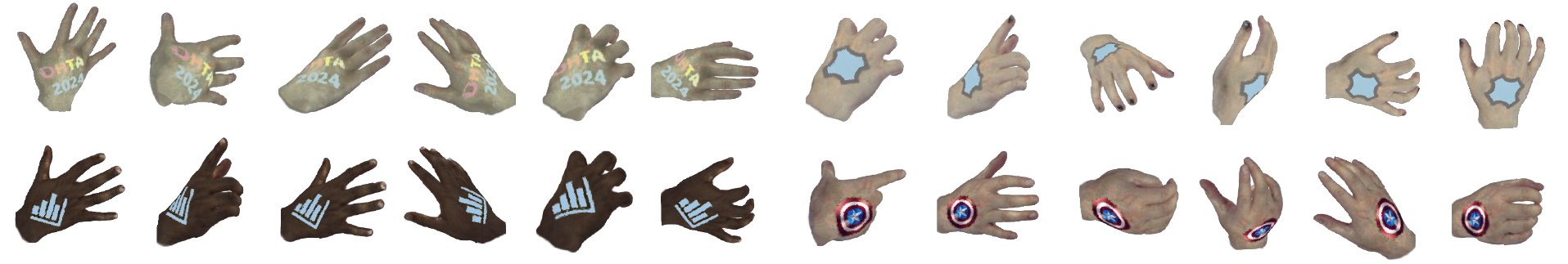}
    \caption{Appearance editing results upon one-shot avatars. }
    \label{fig:supp-editing}
\end{figure*}

\begin{figure*}
    \centering
    \vspace{-0.3cm}
    \includegraphics[width=1.00\textwidth]{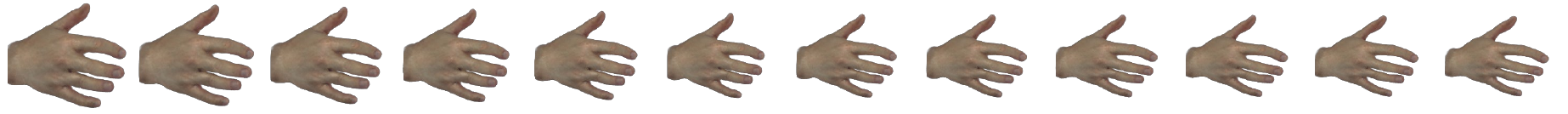}
    \caption{Shape editing results upon one-shot avatars. }
    \label{fig:supp-shape-editing}
\end{figure*}

We provide more results for text-to-avatar in \cref{fig:supp-t2a-result}, appearance editing in \cref{fig:supp-editing}, and shape editing in \cref{fig:supp-shape-editing}.
The animatable hand avatars from text prompts and appearance editing justify OHTA's capability to capture the highly complex details for high-fidelity modeling.
Moreover, the shape editing further validates that the mesh-guided design can fully transfer the texture prior to the novel hand shape, which is essential for the one-shot reconstruction.

\subsection{Video Demo}
We also provide additional qualitative results in the attached video.

\section{Limitations and Failure Cases}
In this section, we outline the limitations of our method to provide a more complete sense of the work's scope.

\begin{figure}[ht!]
    \centering
    \includegraphics[width=\columnwidth]{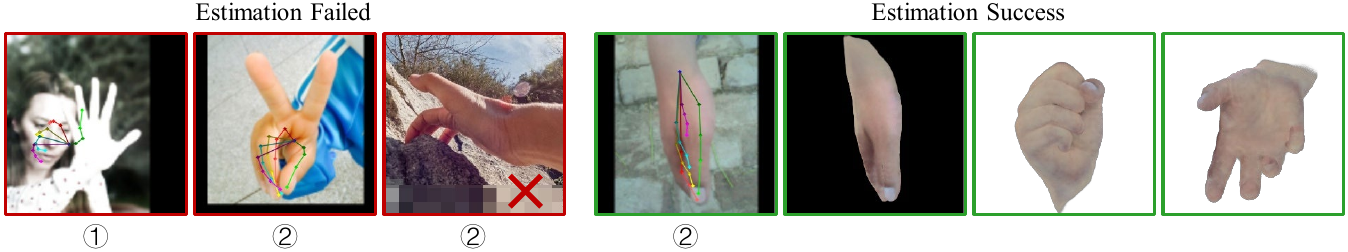}
    \caption{Failure cases of the estimator (left). With successful estimations, OHTA creates hand avatars (right). The images come from the Onehand10K~\cite{wang2018mask} dataset or the internet.}
    \label{fig:failure_estimator}
\end{figure}

OHTA relies on pose estimations for creating avatars.
For some challenging cases as shown in \cref{fig:failure_estimator}, \eg, \textcircled{\raisebox{-0.9pt}{1}} challenging lighting conditions and \textcircled{\raisebox{-0.9pt}{2}} skewed hand images, OHTA may fail because the estimated poses have significant errors or completely fail.
However, if the pose estimation is reliable, OHTA has the robustness to produce results.

\begin{figure*}
    \centering
    \vspace{-0.3cm}
    \includegraphics[width=1.00\textwidth]{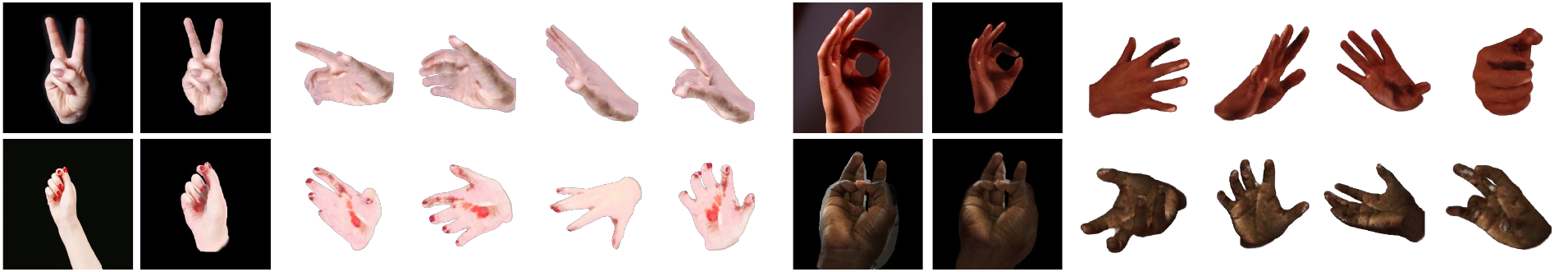}
    \caption{Failure cases due to notably uneven lighting (in the first row) or highly inaccurate pose estimation results (in the second row).}
    \label{fig:supp-failure-case}
\end{figure*}

For 1) notably uneven lighting, and 2) some challenging poses with inaccurate estimations, OHTA may still fail.
OHTA cannot resolve notably uneven lighting since it cannot detect the hand shadows on the input image.
As shown in the first row of \cref{fig:supp-failure-case}, using images with severe shadows for one-shot reconstruction often results in undesirable shadows on avatars and unnatural transitions between observed and unobserved parts.
OHTA relies on the estimated pose results for geometry modeling, which forms the basis for texture modeling.
Therefore, poor estimation results lead to inferior texture modeling, especially for challenging poses.
Even though OHTA is robust to inaccurate estimations to some extent, as shown in the results for in-the-wild images, due to the learned hand priors, it is still incapable of addressing highly inaccurate estimations for some challenging poses.
The second row of \cref{fig:supp-failure-case} illustrates the texture misalignment and artifacts caused by those challenging poses with inaccurate estimations.

Another limitation relates to the number of fingers.
Since the estimator and OHTA both utilize a hand parametric model (\eg MANO) with only five fingers.
It is hard to resolve hands with more or less than five fingers.
We believe that exploring the use of non-parametric representations could help address this challenge.

\clearpage

\clearpage

\clearpage

\clearpage

\setlength{\abovedisplayskip}{0pt}
\setlength{\belowdisplayskip}{0pt}
\setlength{\abovedisplayshortskip}{0pt}
\setlength{\belowdisplayshortskip}{0pt}
\setlength\abovecaptionskip{3pt plus 2pt minus 1pt}

\clearpage

{
    \small
    \bibliographystyle{ieeenat_fullname}
    \bibliography{main}
}

\end{document}